%% file: arxiv.tex
\PassOptionsToPackage{table}{xcolor}
\documentclass[10pt,twocolumn,letterpaper]{article}

\usepackage[pagenumbers]{cvpr}

\usepackage[T1]{fontenc}

\def\confName{CVPR}
\def\confYear{2027}

\input{math_commands.tex}

\usepackage{url}
\usepackage{amsmath}
\usepackage{amssymb}
\usepackage{booktabs}
\usepackage{listings}
\usepackage[most]{tcolorbox}
\usepackage{multirow}
\usepackage{makecell}
\usepackage{array}
\usepackage{tabularx}
\usepackage{threeparttable}
\usepackage[nopatch=eqnum]{microtype}
\usepackage{placeins}
\usepackage{float}
\usepackage{caption}
\usepackage{algorithm}
\usepackage{algpseudocode}
\usepackage{enumitem}
\usepackage{balance}
\definecolor{cvprblue}{rgb}{0.21,0.49,0.74}
\usepackage[pagebackref,breaklinks,colorlinks,allcolors=cvprblue]{hyperref}
\usepackage{fvextra}
\definecolor{stormblue}{HTML}{2F66E8}
\definecolor{stormhead}{HTML}{E7EEF8}
\definecolor{stormdark}{HTML}{15223A}
\definecolor{stormgray}{HTML}{5D6675}
\newcommand{\tabtitle}[1]{\textbf{#1}}
\newcommand{\tabhead}[1]{\textbf{#1}}
\newcommand{\best}[1]{\textbf{#1}}
\newcommand{\second}[1]{\underline{#1}}
\newcommand{\KA}{\mathrm{KA}}
\newcommand{\AR}{\mathrm{AR}}
\newcommand{\HM}{H}
\newcommand{\bench}{RevisitQA}
\renewcommand{\arraystretch}{1.08}
\newcommand{\showprojecturl}{}

\title{RevisitQA: Short Videos Are Not Easy under Dense Revisits}

\author{%
Siru Zhong$^{1,2}$,
Shenghan Tan$^{3,2}$,
Rihong Yan$^{4,2}$,
Xiaohui Lv$^{2}$,
Yuzheng Zhuang$^{2}$,\\
Shuai Tao$^{2,*}$,
Wulong Liu$^{2}$,
Haohuan Fu$^{5}$,
Yuxuan Liang$^{1,*}$\\
$^{1}$HKUST (GZ) \enspace
$^{2}$Beta Infinity \enspace
$^{3}$Beihang University \enspace
$^{4}$UESTC \enspace
$^{5}$Tsinghua University
}

\begin{document}
\raggedbottom

\maketitle

\input{contents/00_abs}
\input{contents/01_intro_arxiv}
\input{contents/02_related_work_arxiv}
\input{contents/03_method}
\input{contents/04_experiments}
\input{contents/05_conclusion}

\section*{LLM Usage Statement}
We used Doubao Seed~2.1~Pro, Claude~Opus~5.5, GPT-6.1~Sol, and open video VLMs for episode timelines, question drafting and distractors, text debiasing, text-prior filtering, label cross-checks, and manuscript edits. After reviewing all AI-assisted work, we take full responsibility for this paper; tool notes appear in the appendix.

\bibliography{storm}
\bibliographystyle{ieeenat_fullname}

\appendix
\renewcommand{\thefigure}{S\arabic{figure}}
\renewcommand{\thetable}{S\arabic{table}}
\renewcommand{\theequation}{S\arabic{equation}}
\setcounter{figure}{0}
\setcounter{table}{0}
\setcounter{equation}{0}

\input{contents/appendix_construction}

\end{document}

%% file: math_commands.tex
\usepackage{amsmath,amsfonts,bm}

\def\eqref#1{(\ref{#1})}

\def\1{\bm{1}}

\DeclareMathAlphabet{\mathsfit}{\encodingdefault}{\sfdefault}{m}{sl}
\SetMathAlphabet{\mathsfit}{bold}{\encodingdefault}{\sfdefault}{bx}{n}



%% file: contents/00_abs.tex
\begin{abstract}
Video question answering benchmarks have increasingly focused on longer recordings, yet short videos can also be challenging when scenes are revisited frequently. As a camera repeatedly leaves and returns to the same workspace, objects may change out of view, making earlier observations outdated or misleading. Understanding such videos requires models to track evolving states across repeated visits rather than simply retrieve past observations. To study this challenge, we introduce \bench{}, a benchmark containing 2,577 questions across 622 episodes with a median duration of 44.8 seconds, drawn from five real-world egocentric domains and two simulated environments. \bench{} covers six question types and evaluates models at specific query times, requiring them to answer from the observed history or abstain when it is insufficient. Across 15 offline and seven streaming video LLMs, the median model correctly answers only 55.5\% of answerable questions and abstains on just 30.3\% of unanswerable ones. Removing earlier frames further reveals that historical observations help answer questions but can also make models less likely to abstain when answers remain unresolved. These findings highlight dense episodic revisits as an important and unresolved challenge for video question answering.
\ifdefined\showprojecturl
Code is available at \url{https://github.com/siruzhong/RevisitQA}.
\fi
\end{abstract}

%% file: contents/01_intro_arxiv.tex
\section{Introduction}
\label{sec:intro}

Multimodal large language models (MLLMs) have expanded video question answering (Video-QA) from static clip recognition to temporal reasoning~\citep{li2024mvbench,fu2025videomme}. Conventional Video-QA evaluates a model after the entire video is observed. Interactive settings instead ask at specific timestamps, so the model sees only the causal history up to query time~\citep{yang2025svbench,huang2025ovbench}. This setting becomes particularly challenging when a camera repeatedly returns to a scene whose state may have changed.

Existing benchmarks study retrieval, state evolution, and selective answering from different perspectives. Long-video benchmarks extend the horizon to minute- and hour-long recordings for long-range retrieval~\citep{mangalam2023egoschema,wu2024longvideobench,zhou2024mlvu,wang2024lvbench}, and streaming and memory-augmented models are built to compress those histories~\citep{chen2024videollmonline,qian2025dispider,wang2025streambridge,he2024malmm}. State-tracking suites follow how answers evolve over time~\citep{gao2021envqa,xun2025rtvbench,liu2026svcbench,forte2026egostream}. Selective VQA and video abstention benchmarks induce or score unanswerability~\citep{whitehead2022reliable,dancette2023selective,yoon2025canvideollms,wu2026abstaineqa,yu2026virtuebench,pramono2026trapsbench}, while StreamReady~\citep{azad2026streamready} asks when enough evidence has arrived.

\begin{figure*}[t]
  \centering
  \includegraphics[width=\textwidth]{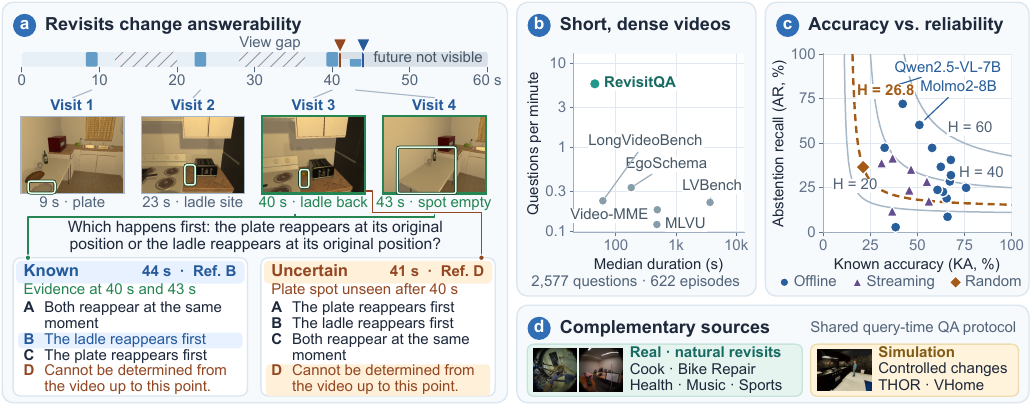}
  \caption{\textbf{Overview of \bench{}.}
    (a)~Revisits change answerability: Uncertain at 41\,s before the plate site is re-seen, Known at 44\,s after an empty view.
    (b)~Episodes are short and densely annotated relative to long-form Video-QA.
    (c)~Mean $\KA$ vs.\ $\AR$ with $\HM$ contours; diamond marks random abstention ($\HM=26.8$).
    (d)~Five real domains plus THOR and VHome under a shared query-time Known/Uncertain protocol.}
  \label{fig:teaser}
  \vspace{-1em}
\end{figure*}

\looseness=-1
Short videos with \emph{dense episodic revisits} pose a distinct challenge. When an egocentric camera or agent repeatedly leaves and returns to the same workspace~\citep{perrett2025hd,grauman2024egoexo4d}, episodes remain short ($\sim$1\,min) yet may contain frequent state changes (Figure~\ref{fig:teaser}b). Across visits, objects may be relocated, altered, or occluded, and earlier observations become \emph{stale}: a mechanic holds a lubricant bottle at 16\,s but a red cloth at 23\,s, yet Qwen3-VL-4B still answers with the bottle. Other changes occur out of view, leaving the latest state unresolved. Answerability can also change with new observations: in Figure~\ref{fig:teaser}a, the same question is \textit{Uncertain} at 41\,s before the plate site is re-seen, then \textit{Known} at 44\,s after an empty view.

To study this challenge, we introduce \textbf{\bench{}} with 2,577 questions across 622 short episodes (44.8-s median) sampled at 1~FPS. We compile revisits to the same functional region in Cook from HD-EPIC~\citep{perrett2025hd} and four Ego-Exo4D domains~\citep{grauman2024egoexo4d} into short real-world revisit episodes. Continuous camera rollouts in two simulated environments, THOR (AI2-THOR~\citep{kolve2017ai2thor}) and VHome (VirtualHome~\citep{puig2018virtualhome}), provide controlled scenarios (Figure~\ref{fig:teaser}d). Questions span six semantic types and are labeled by accumulated change intensity and query-time answerability: \textit{Known} when the observed prefix supports an answer, and \textit{Uncertain} when it does not. Annotations record evidence spans, question type, and uncertainty source. Every item offers three options plus a shared abstention. We score Known accuracy ($\KA$), abstention recall ($\AR$), and their harmonic mean $\HM$ (Figure~\ref{fig:teaser}c).

Across 22 video LLMs, including 15 offline and seven streaming models, ranking by $\KA$ is nearly unrelated to ranking by $\HM$ ($\tau=0.06$): the most accurate model, Qwen3.5-9B ($\KA$ 75.8\%), abstains on only 25.0\% of Uncertain questions. Models are less likely to abstain when observations are missing (26.6\% $\AR$) than when frames are visually degraded (62.3\% $\AR$). Removing earlier frames further reduces accuracy on questions requiring older observations while increasing abstention on Uncertain questions. These findings highlight the difficulty of maintaining and using history across dense revisits. Our main contributions are:

\begin{itemize}[leftmargin=*]
  \item \textbf{A dense-revisit video benchmark.} \bench{} combines physical transitions and episodic revisits in short episodes across five real domains and two simulators, with code and prompts provided for reproduction and extension.
  \item \textbf{Query-time evaluation across revisits.} We label Known or Uncertain from the observed prefix, score $\KA$ and $\AR$, and rank by $\HM$. Text-only, black-frame, and forced-choice controls measure how answers depend on the video.
  \item \textbf{Diagnostics across 22 video LLMs.} Breakdowns by change intensity, evidence window, question type, and uncertainty source show where models fail under scene changes, stale evidence, and missing observations.
\end{itemize}

%% file: contents/02_related_work_arxiv.tex
\section{Preliminaries and Related Work}
\label{sec:prelim}

\subsection{Problem Formulation}
\label{sec:protocol}

We evaluate interactive Video-QA on a causal prefix. At query time $t_q$, the model sees only $V_{\le t_q}$ and must either select a substantive answer or abstain ($\bot$) when the prefix is insufficient. Under dense episodic revisits, the camera repeatedly leaves and returns to the same regions: earlier observations can become \emph{stale evidence} after a state change, and the latest transition may remain unobserved. Whether the question is answerable depends on the observation history up to $t_q$; we label queries \emph{Known} or \emph{Uncertain} and score Known accuracy ($\KA$) with abstention recall ($\AR$). Models are ranked by their harmonic mean $\HM=2\,\KA\cdot\AR/(\KA+\AR)$.

\subsection{Related Work}
\label{sec:related}

\input{tables/tab_compare}

\begin{figure*}[t]
  \centering
  \includegraphics[width=\textwidth]{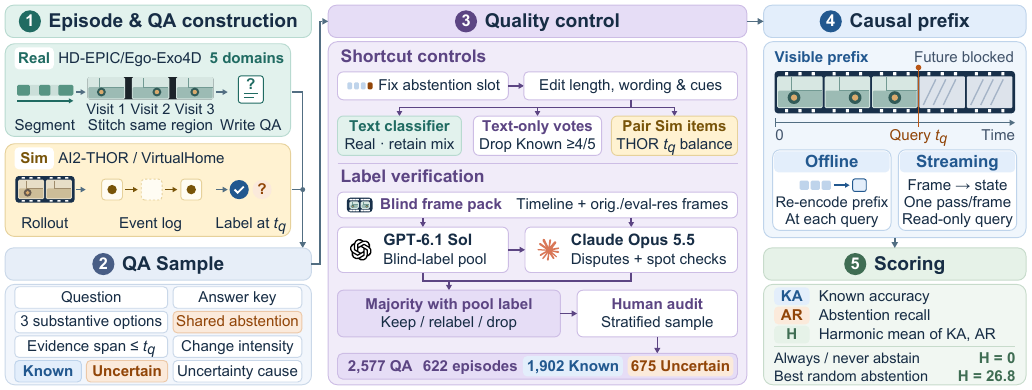}
  \caption{\textbf{Construction, quality control, and evaluation of \bench{}.} (1)~Real-world episodes stitch revisits to one region; simulated episodes use camera rollouts labeled at $t_q$. (2)~Each item has three options plus abstention, evidence spans $W_i\subseteq[0,t_q]$, and diagnostic labels. (3)~Shortcut filters, conditional verification, and a human audit yield 2,577 items. (4)~Offline models re-encode the prefix per query; streaming models update once per frame. (5)~Scoring uses $\KA$, $\AR$, and $\HM$.}
  \label{fig:framework}
  \vspace{-1em}
\end{figure*}

\paragraph{Video-QA Benchmarks.}
Prior Video-QA suites scale duration or move evaluation onto causal streams. Long-video benchmarks such as EgoSchema~\citep{mangalam2023egoschema}, Video-MME~\citep{fu2025videomme}, LongVideoBench~\citep{wu2024longvideobench}, MLVU~\citep{zhou2024mlvu}, and LVBench~\citep{wang2024lvbench} stress long-range retrieval, while streaming suites such as StreamingBench~\citep{lin2024streamingbench}, SVBench~\citep{yang2025svbench}, OVBench~\citep{huang2025ovbench}, and OVO-Bench~\citep{li2025ovobench} ask at timestamps under a growing prefix. Short-clip~\citep{li2024mvbench,plizzari2025egotempo} and egocentric-memory~\citep{grauman2022ego4d,majumdar2024openeqa} settings recover content from a recording; they do not target returns to a workspace that changed out of view. \bench{} (Table~\ref{tab:compare}) targets dense episodic revisits within short videos, where repeated out-of-view changes make earlier observations stale and leave the latest state unobserved at query time. Difficulty can therefore arise from revisit-driven state changes, even in short videos, and each query must resolve the latest observed state from the causal prefix alone.
\vspace{-1em}

\paragraph{State Tracking and Abstention.}
Env-QA~\citep{gao2021envqa}, RTV-Bench~\citep{xun2025rtvbench}, SVCBench~\citep{liu2026svcbench}, and EgoStream~\citep{forte2026egostream} follow how answers evolve over time. Selective VQA~\citep{whitehead2022reliable,dancette2023selective} couples answering with abstention, and recent studies show that models rarely decline unanswerable questions~\citep{yoon2025canvideollms,kirichenko2025abstentionbench,lei2026egosat}, while video abstention suites often \emph{induce} unanswerability by editing questions or inputs~\citep{wu2026abstaineqa,yu2026virtuebench,pramono2026trapsbench,li2025ovobench}. StreamReady~\citep{azad2026streamready} primarily scores \emph{when} to answer against evidence windows and probes unanswerability with separate counterfactual questions. Under dense episodic revisits, a question answerable on one visit can become unanswerable after an unobserved change and answerable again upon a revisit, so \bench{} keeps the selective protocol and derives Known/Uncertain labels from natural view gaps, scoring both jointly with accumulated change intensity.
\vspace{-1em}

\paragraph{Offline and Streaming Video LLMs.}
Offline video LLMs compress or reason over a fixed clip~\citep{llavavl2024llavanextvideo,wang2025internvideo25,zhang2025videollama3,yao2025minicpmv45,wang2025internvl35,chen2025eagle25,qwenteam2026qwen35,vteam2025glm41v,bai2025qwen25vl,bai2025qwen3vl,clark2026molmo2}, while streaming and memory-augmented models update a compact state as frames arrive~\citep{song2024moviechat,he2024malmm,zhang2025flashvstream,xiong2025streamchat,azad2025hierarq,di2025rekv,yao2025timechatonline,chen2024videollmonline,chen2025livecc,qian2025dispider,wang2025streambridge}. These designs mainly target long histories. \bench{} evaluates both on fine-grained state tracking and selective abstention across observation gaps in short episodes under one query-time protocol (\S\ref{sec:rq1}).

%% file: tables/tab_compare.tex
\begin{table}[t]
\centering
\providecommand{\cmark}{\textcolor[HTML]{22978A}{\ensuremath{\checkmark}}}
\providecommand{\xmark}{\textcolor[HTML]{9DAFC1}{\ensuremath{\times}}}
\providecommand{\pmark}{\textcolor[HTML]{A35B17}{\ensuremath{\circ}}}
\caption{\textbf{Comparison with representative Video-QA benchmarks.} \textit{Causal}: only frames up to the query time $t_q$ are visible when answering. \textit{Revisit}: evaluation targets returns to regions that changed out of view. \textit{Ans.\,$(t_q)$}: the answer or answerability depends on $t_q$. \textit{Nat.}: missing evidence arises from out-of-view workspace intervals or line-of-sight occlusions. \textit{Abst.}: abstention is scored. \textit{Cause}: Uncertain items carry cause labels. Marks: \cmark{}~yes, \pmark{}~partial, \xmark{}~not targeted by the benchmark.}
\label{tab:compare}
\scriptsize
\setlength{\tabcolsep}{1.6pt}
\renewcommand{\arraystretch}{1.05}
\begin{tabularx}{\columnwidth}{@{}l*{10}{>{\centering\arraybackslash}X}>{\columncolor{stormhead!60}\centering\arraybackslash}X@{}}
\toprule
\rowcolor{stormhead!55}
 & \textbf{ES} & \textbf{VM} & \textbf{SB} & \textbf{OVO} & \textbf{RTV} & \textbf{SVC} & \textbf{ESt} & \textbf{PR} & \textbf{VB} & \textbf{TB} & \cellcolor{stormhead!82!stormblue}\textbf{Ours} \\
\midrule
Causal & \xmark & \xmark & \pmark & \pmark & \pmark & \cmark & \cmark & \xmark & \xmark & \xmark & \cmark \\
Revisit & \xmark & \xmark & \xmark & \xmark & \xmark & \xmark & \xmark & \xmark & \xmark & \xmark & \cmark \\
Ans.\,$(t_q)$ & \xmark & \xmark & \cmark & \cmark & \cmark & \cmark & \pmark & \cmark & \xmark & \xmark & \cmark \\
Nat. & \xmark & \xmark & \xmark & \xmark & \xmark & \xmark & \xmark & \xmark & \xmark & \xmark & \cmark \\
Abst. & \xmark & \xmark & \xmark & \pmark & \xmark & \xmark & \xmark & \pmark & \cmark & \cmark & \cmark \\
Cause & \xmark & \xmark & \xmark & \xmark & \xmark & \xmark & \xmark & \xmark & \xmark & \cmark & \cmark \\
\bottomrule
\end{tabularx}\\[3pt]
\begin{minipage}{\columnwidth}
\scriptsize
ES:~EgoSchema~\citeyearpar{mangalam2023egoschema};
VM:~Video-MME~\citeyearpar{fu2025videomme};
SB:~StreamingBench~\citeyearpar{lin2024streamingbench};
OVO:~OVO-Bench~\citeyearpar{li2025ovobench};
RTV:~RTV-Bench~\citeyearpar{xun2025rtvbench};
SVC:~SVCBench~\citeyearpar{liu2026svcbench};
ESt:~EgoStream~\citeyearpar{forte2026egostream};
PR:~StreamReady~\citeyearpar{azad2026streamready};
VB:~VirtueBench~\citeyearpar{yu2026virtuebench};
TB:~TRAPSBench~\citeyearpar{pramono2026trapsbench};
Ours:~\bench{}.
\end{minipage}
\vspace{-1em}
\end{table}

%% file: contents/03_method.tex
\section{\texorpdfstring{\bench{}}{RevisitQA}}
\label{sec:method}

\bench{} follows the pipeline in Figure~\ref{fig:framework}. Real episodes stitch revisits to one region, and simulation episodes use continuous rollouts labeled at $t_q$ (Step~1). Each item carries three options plus a shared abstention option, evidence spans $W_i\subseteq[0,t_q]$, and diagnostic labels (Step~2). Shortcut filters, conditional two-model verification, and a stratified human audit yield 2,577 questions across 622 1~FPS episodes (Step~3). Offline models re-encode the causal prefix per query, while streaming models update once per frame without writing answers back (Step~4). We score $\KA$, $\AR$, and $\HM$ under a shared causal-prefix protocol (Step~5).

\subsection{Episode Construction}
\label{sec:construction}

\bench{} combines five real-world egocentric domains with two continuous simulated camera rollouts. Both tracks use the same 1~FPS causal-prefix protocol, Known/Uncertain labels, and change-intensity stratification.
\vspace{-1.5em}

\paragraph{Real-world egocentric episodes.}
The real-world track spans Cook from HD-EPIC~\citep{perrett2025hd} and Bike Repair, Health, Music, and Sports from Ego-Exo4D~\citep{grauman2024egoexo4d}. In raw egocentric recordings, returns to the same workspace are separated by minutes of activity elsewhere. We retain visits to a shared functional region and compile them into revisit episodes (Figure~\ref{fig:framework}, Step~1). Source videos sampled at 1~FPS are segmented by a VLM into region-activity timelines that record functional zones, manipulations, and state transitions (relocation, transformation, replacement). An LLM planner clusters segments by functional region and assembles 3--10 ordered visits into $\sim$45\,s episodes with 0.2\,s black transitions at inter-visit cuts and 1.5$\times$ playback before 1~FPS sampling, so successive returns are separated by out-of-view gaps.
\vspace{-1em}

\paragraph{Continuous simulation control.}
The simulation track provides continuous simulated camera rollouts in THOR (AI2-THOR~\citep{kolve2017ai2thor}) and VHome (VirtualHome~\citep{puig2018virtualhome}) across Kitchen, Living Room, Bedroom, and Bathroom scenes. Observation gaps arise from camera trajectories and line-of-sight occlusions. Seeded camera routes create view gaps around ten scheduled events: disappearance and reappearance interventions in THOR, and multi-agent open/close, switch, and pickup/placement actions in VHome. Execution traces and instance masks link each state change to visible before and after frames. VHome requires two consecutive 1~FPS frames for a visibility transition and uses RGB endpoint review for physical-state questions. Query times and evidence spans satisfy $W_i\subseteq[0,t_q]$, so seeing a location empty can resolve a visibility query while an unobserved state change stays Uncertain. Deterministic builders instantiate six families from VLM-polished templates with shared abstention.

\subsection{Question Taxonomy and Diagnostic Axes}
\label{sec:taxonomy}

Each question is posed at a query timestamp $t_q$ with grounded evidence spans $W_i \subseteq [0, t_q]$ and four options $\mathcal{A}=\{A,B,C,D\}$: three substantive candidates $\mathcal{Y}_q$ (shuffled deterministically per item) and the shared abstention option $\bot$ (\emph{``Cannot be determined from the video up to this point.''}). The abstention option keeps the same wording and position on every item (Figure~\ref{fig:framework}, Step~2). Questions span six semantic types: factual retrieval, current state, state change, object tracking, temporal reasoning, and history aggregation. Every question is also stratified by accumulated change intensity and Known/Uncertain answerability.
\vspace{-0.5em}

\paragraph{Accumulated change intensity.}
Each question receives a track-specific ordinal proxy $c_i \in \{1, \dots, 10\}$ at $t_q$, binned as Low ($B_1=[1,3]$), Medium ($B_2=[4,6]$), and High ($B_3=[7,10]$). In the real-world track, $c_i$ counts initiated revisit intervals to the queried functional region up to $t_q$. In the simulation track, it counts scheduled event intervals whose onset is no later than $t_q$. Higher $c_i$ indicates more accumulated revisit or event activity prior to the query under the shared Low/Medium/High bins.
\vspace{-0.5em}

\paragraph{Query-time answerability.}
Following Section~\ref{sec:protocol}, each query is labeled Known when $V_{\le t_q}$ supports a unique substantive answer $a_i \in \mathcal{Y}_q$, and Uncertain when the observed prefix does not support a unique substantive answer ($a_i = \bot$). To diagnose abstention failures, each Uncertain item carries one or more of six uncertainty sources: (1)~\textit{missing observation}, (2)~\textit{partial observation or occlusion}, (3)~\textit{ambiguous evidence}, (4)~\textit{low visual quality}, (5)~\textit{multiple plausible candidates}, and (6)~\textit{ambiguous attribute} for an unresolved queried property.

\vspace{0.5em}
\subsection{Quality Control and Label Verification}
\label{sec:qc}

We combine construction-time shortcut filtering with conditional label verification and a stratified human audit so that released questions depend on visual evidence up to $t_q$ under the shared 1~FPS causal-prefix protocol (Figure~\ref{fig:framework}, Step~3).
\vspace{-0.5em}

\paragraph{Controls against non-visual shortcuts.}
Options are equalized in length, syntax, and specificity so distractors differ in visually grounded properties. In the real-world track, an episode-grouped cross-validated text classifier filters text-predictable items while preserving their distribution. In THOR, paired Known and Uncertain items share templates and option sets and are balanced across query time. We remove Known items answered without video by at least four of five filter VLMs (Qwen2.5-VL-3B/7B, Qwen3-VL-8B, InternVL3.5-8B, and MiniCPM-V-4.5). Residual option-length checks and restore-filter sensitivity are in the supplement. These models also serve as the \S\ref{sec:rq4} controls (options-only, text-only, black-frame, shuffled-frame, forced-choice).
\vspace{-2em}

\paragraph{Label verification.}
All items follow a unified Known/Uncertain rubric that accounts for legibility at evaluation resolution. Let the pool label denote the construction-time answerability label. GPT-6.1 Sol labels real-world, VHome, and paired THOR candidates from a frame package truncated at $t_q$; other THOR Known builders use simulator traces and instance masks. Claude Opus 5.5 labels all disagreements with the pool label, plus a 15\% sample of agreeing items. Final labels follow a majority vote over the pool label and the available model judgments; unresolved items are discarded. A stratified audit of 141 items by two independent annotators yielded 84.4\% inter-annotator agreement (Cohen's $\kappa=0.688$), with 85.1\% and 83.7\% agreement with the release labels (supplement).

\subsection{Evaluation Metrics}
\label{sec:scoring}

Given $V_{\le t_q}$, the question, and the four options $\mathcal{A}$, a model predicts a single choice $\hat{a}_i$ under the Step~4 protocol (Figure~\ref{fig:framework}). Unparseable outputs are counted as incorrect answers.
Let $\mathcal{K}$ and $\mathcal{U}$ denote Known and Uncertain items. We score answering and abstaining separately as
\begin{equation}
\KA = \frac{1}{|\mathcal{K}|}\sum_{i\in\mathcal{K}}\mathbb{1}[\hat{a}_i{=}a_i],
\AR = \frac{1}{|\mathcal{U}|}\sum_{i\in\mathcal{U}}\mathbb{1}[\hat{a}_i{=}\bot],
\end{equation}
and rank by $\HM = \frac{2\,\KA\cdot\AR}{\KA+\AR}$, following generalized zero-shot learning~\citep{xian2017zero}.
Overall $\HM$ is the unweighted average of the seven per-subset harmonic means.
Always-abstain and never-abstain yield $\HM=0$; optimal random abstention gives the chance baseline $\HM=26.8$.
We stratify these scores by change-intensity bin, evidence window, question type, and uncertainty source, and report wrong abstention on Known items (FAR) separately from wrong answers; $1-\AR$ is the overconfidence rate on Uncertain items.
All 95\% confidence intervals use 1,000 episode-clustered bootstrap resamples.

%% file: contents/04_experiments.tex
\section{Experiments}
\label{sec:experiments}

In this section, we evaluate 22 video LLMs on \bench{} along four research questions.
\begin{itemize}[leftmargin=*]
  \item \textbf{RQ1} examines performance on short episodes with dense revisits~(Sec.~\ref{sec:rq1}).
  \item \textbf{RQ2} tests whether the metric or input protocol changes the conclusions~(Sec.~\ref{sec:rq2}).
  \item \textbf{RQ3} analyzes failure patterns across change intensity, uncertainty sources, and question types~(Sec.~\ref{sec:rq3}).
  \item \textbf{RQ4} tests whether answers rely on the video and on valid rather than stale history~(Sec.~\ref{sec:rq4}).
\end{itemize}

\subsection{Experimental Setup}
\label{sec:setup}

\begin{figure}[b]
  \centering
  \includegraphics[width=\linewidth]{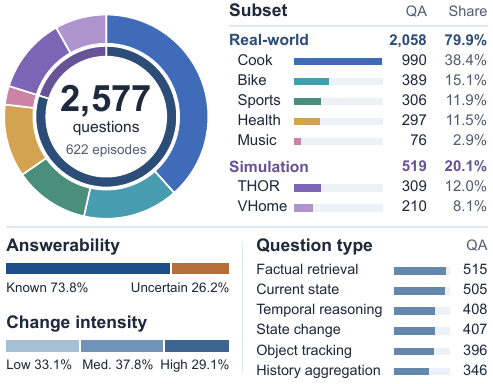}
  \caption{\textbf{Composition of \bench{}.} Inner/outer rings show real--sim and subset shares; lower charts show answerability, change intensity, and question types over the 2,577-item release.}
  \label{fig:dataset}
\end{figure}

\paragraph{Dataset statistics and protocol.}
\bench{} consists of seven subsets covering 2,577 questions across 622 episodes and 7.56\,h of video across five real domains and two simulators, as shown in Figure~\ref{fig:dataset}. Real-world items make up 79.9\% of the bench and simulation 20.1\%. Of these questions, 26.2\% are Uncertain, change intensity splits across Low, Medium, and High (33.1\%/37.8\%/29.1\%), and 42.1\% need evidence that starts before the latest 16 frames, ranging from 29.7\% in Sports to 57.9\% in Music. The same figure also breaks items down by the six semantic question types. All subsets are evaluated zero-shot under the 1~FPS causal-prefix protocol ($V_{\le t_q}$, up to 200,704 pixels per frame), with a full-video control ($V$). We report unweighted means of $\KA$, $\AR$, and $\HM$ over the seven subsets with 1,000 episode-clustered bootstrap CIs, and compare model rankings across metrics on all seven domains for every result reported below.

\paragraph{Evaluated models.}
We evaluate 15 offline video LLMs from 11 families: Qwen2.5-VL (3B/7B)~\citep{bai2025qwen25vl}, Qwen3-VL (4B/8B/32B)~\citep{bai2025qwen3vl}, Qwen3.5 (4B/9B)~\citep{qwenteam2026qwen35}, InternVL3.5-8B~\citep{wang2025internvl35}, Molmo2-8B~\citep{clark2026molmo2}, MiniCPM-V~4.5~\citep{yao2025minicpmv45}, InternVideo2.5-8B~\citep{wang2025internvideo25}, Eagle2.5-8B~\citep{chen2025eagle25}, VideoLLaMA3-7B~\citep{zhang2025videollama3}, LLaVA-NeXT-Video-7B-DPO~\citep{llavavl2024llavanextvideo}, and GLM-4.1V-9B-Thinking~\citep{vteam2025glm41v}. We also evaluate 7 streaming models with different state mechanisms: compressed memory banks (Flash-VStream~\citep{zhang2025flashvstream}, MovieChat~\citep{song2024moviechat}), hierarchical memory retrieval (StreamChat~\citep{xiong2025streamchat}), KV-cache retrieval (ReKV~\citep{di2025rekv}), differential token dropping (TimeChat-Online~\citep{yao2025timechatonline}), interleaved video--text training (LiveCC~\citep{chen2025livecc}), and a response-timing module (Dispider~\citep{qian2025dispider}). Each ingests 1\,FPS frames once per episode with its native memory. Unless noted, the five control models are the Sec.~\ref{sec:qc} filter VLMs (Qwen2.5-VL-3B/7B, Qwen3-VL-8B, InternVL3.5-8B, MiniCPM-V-4.5), used in the RQ2--RQ4 controls.

\subsection{RQ1: Whole-Bench Evaluation}
\label{sec:rq1}

\input{tables/tab_real_results}

\begin{figure*}[t]
  \centering
  \includegraphics[width=\linewidth]{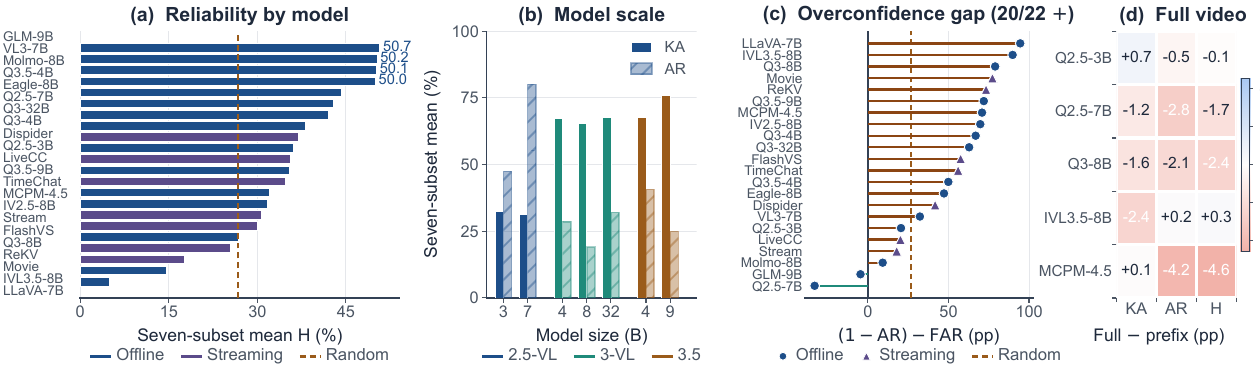}
  \caption{\textbf{Whole-bench results.} (a)~All 22 models by seven-subset mean $\HM$ (offline vs.\ streaming); dashed line is random abstention ($\HM=26.8$). (b)~$\KA$ (filled) and $\AR$ (hatched) by size within three Qwen families. (c)~Overconfidence gap $(1-\AR)-\mathrm{FAR}$ (pp); positive means missed abstention on Uncertain exceeds FAR on Known. (d)~Full-video minus causal-prefix deltas (heatmap) for five controls.}
  \label{fig:results}
  \vspace{-1em}
\end{figure*}

\paragraph{Overall performance.}
We evaluate how well current video models handle short episodes with dense revisits. Across the 22 models, median $\KA$ is 55.5\%, while median $\AR$ is only 30.3\%; 19 of 22 abstain on fewer than half of the Uncertain items (Table~\ref{tab:real-results}). Several fall below the random-abstention baseline of $\HM=26.8$ (Figure~\ref{fig:results}a), and for most models the main failure is answering Uncertain items instead of abstaining (Figure~\ref{fig:results}c). The top four models all reach $\HM\approx 50$ but with opposite profiles: GLM-4.1V-9B-Thinking and Molmo2-8B lean toward abstaining, while VideoLLaMA3-7B and Qwen3.5-4B lean toward answering. Increasing model size does not consistently improve both $\KA$ and $\AR$ (Figure~\ref{fig:results}b). For example, Qwen3.5-9B improves $\KA$ over its 4B counterpart by 8.1 points while reducing $\AR$ by 15.6 points; other Qwen families exhibit different scaling patterns.

\paragraph{Offline vs.\ streaming architectures.}
To test whether keeping a running memory helps under dense revisits, we compare offline and streaming models on the same queries. Streaming models lag offline ones on both $\KA$ and $\AR$: median $\KA$ is 45.6\% vs.\ 62.4\%, and median $\AR$ is 28.6\% vs.\ 32.1\% (Table~\ref{tab:real-results}). Even the best streaming model, Dispider ($\HM=36.9$), falls below the offline median of 38.0, and memory-bank or KV-cache models abstain least; MovieChat and ReKV fall below the random-abstention baseline. Maintaining a streaming memory state therefore does not, by itself, solve dense-revisit QA under the evaluated protocol.

\paragraph{Real-world and simulation subsets.}
Median $\HM$ is 43.4 on the real-world subsets and 18.3 in simulation (Table~\ref{tab:real-results}). On THOR, answering holds up better than abstaining: Qwen3-VL (4B/8B/32B), Qwen3.5-9B, and InternVL3.5-8B keep $\KA$ at 56.5--58.1\% yet abstain on only 0--10.2\% of Uncertain items; median $\AR$ is 7.1\% and five models never abstain correctly (per-model THOR/VHome matrix in the supplementary material). The highest THOR $\HM$ comes from blanket abstention: GLM-4.1V-9B-Thinking abstains on all Uncertain and 51.2\% of Known; Qwen2.5-VL-7B abstains on 89.8\% of Uncertain and 60.0\% of Known items.

\subsection{RQ2: Metric and Protocol Sensitivity}
\label{sec:rq2}

\begin{figure*}[t]
  \centering
  \includegraphics[width=\linewidth]{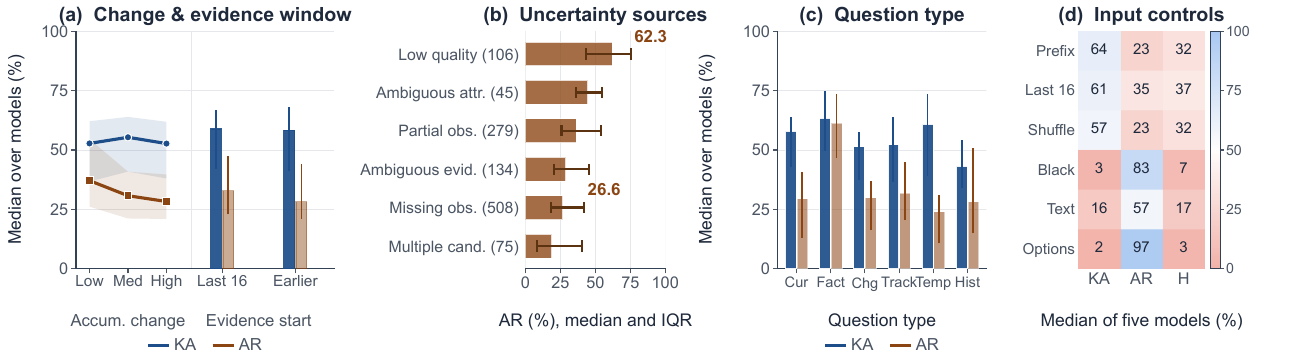}
  \caption{\textbf{Video-QA diagnostics and input controls.}
    (a)~Change-intensity trends (lines, IQR bands) and evidence-window medians (bars).
    (b)~Median $\AR$ by uncertainty source with IQR whiskers ($n$ in parentheses).
    (c)~Median $\KA$/$\AR$ by question type with IQR whiskers.
    (d)~Median $\KA$/$\AR$/$\HM$ heatmap under each input for five controls.
    Per-model tables are in the supplement.}
  \label{fig:diagnostics}
  \vspace{-1em}
\end{figure*}

\paragraph{Rank correlation across metrics.}
Ranking models by $\KA$ is almost unrelated to ranking them by $\HM$ ($\tau=0.06$ across the 22 models), and stronger answering tends to come with weaker abstention. Qwen3.5-9B is the clearest case, with the highest $\KA$ at $75.8\%$ but $\HM$ of only $35.4\%$ because its $\AR$ is $25.0\%$. Overconfidence explains much of the gap: 20 of 22 models miss abstention on Uncertain items more often than they falsely abstain on Known ones (Figure~\ref{fig:results}c).

\paragraph{Causal-prefix vs.\ full-video access.}
We give the five control models the full video, including frames after $t_q$. Access to future frames produces only small changes under the fixed query-time labels: median $\KA$, $\AR$, and $\HM$ each change by 1--2 points, and no model's $\HM$ falls by more than 4.6 (Figure~\ref{fig:results}d; supplement). Four of the five abstain less with future frames, while $\KA$ changes by at most 2.4 points.

\subsection{RQ3: Failure Patterns Across Revisits}
\label{sec:rq3}

\input{tables/tab_abstention_mechanism}

\paragraph{Performance across change intensity.}
We bin questions by accumulated change intensity (Sec.~\ref{sec:taxonomy}). Abstention recall decreases with higher accumulated change intensity, while Known accuracy remains relatively stable: median $\KA$ stays between 52.7\% and 55.4\% from Low to High change, while median $\AR$ drops from 37.1\% to 28.3\% (Figure~\ref{fig:diagnostics}a). On the 42.1\% of items whose evidence starts before the latest 16 frames, Known accuracy stays almost unchanged, but abstention on Uncertain items falls from 32.8\% to 28.2\%.

\paragraph{Sensitivity to uncertainty sources.}
Grouping Uncertain items by uncertainty source (Sec.~\ref{sec:taxonomy}) shows which kinds of insufficiency models detect after leave-and-return gaps. Models abstain when the evidence looks degraded, but often answer when the needed view is missing (Figure~\ref{fig:diagnostics}b): 62.3\% $\AR$ on low-visual-quality items ($n=106$) versus 26.6\% on missing-observation items ($n=508$), 28.7\% on ambiguous-evidence items, and 18.7\% on multiple-candidate items. For example, 16 of 22 models answer earlier microwave-door states that were never shown under occlusion~\mbox{(Fig.~\ref{fig:case-failures}b)}.

\paragraph{Performance by question type.}
Abstention varies sharply with question type (Figure~\ref{fig:diagnostics}c). Only factual retrieval elicits frequent abstention: its $\AR$ is 61.3\%, against 24.0\%--31.7\% on the other types and lowest on temporal reasoning. Median $\KA$ ranges from 43.3\% on history aggregation to 63.3\% on factual retrieval, with state change, object tracking, current state, and temporal reasoning in between. Factual questions often ask for a concrete attribute that is either present in the prefix or not, so models abstain more readily when it is missing. Temporal and state-change questions instead push models to invent an event order across gaps between revisits.

\paragraph{How models abstain.}
The same outputs show that abstention is poorly aligned with error (Table~\ref{tab:abstention-mechanism}). The median model answers 81.2\% of items, yet 48.7\% of those answers are wrong; only 44.3\% of its abstentions fall on Uncertain items (base rate 26.2\%). For rare abstainers such as Qwen3-VL-8B and InternVL3.5-8B, removing the abstention option barely changes $\KA$: they almost never refuse the Known items they would miss under forced choice ($\bot|\times$ is 1.7--2.2). For frequent abstainers such as Qwen2.5-VL, $\KA$ rises sharply without abstention, because they often refuse Known items they would have answered correctly ($\bot|\checkmark$ reaches 26.8--36.9). Rare abstainers refuse too little; frequent ones refuse too broadly. The median model's abstention precision is 44.3 against that 26.2 Uncertain base rate, so the resulting refusals remain only weakly selective.

\input{tables/tab_history_dependence}

\subsection{RQ4: Video Dependence and Stale History}
\label{sec:rq4}

\paragraph{Dependence on the video.}
On the five control models, median $\KA$ falls from 63.5\% with video to 3.5\%, 16.0\%, and 1.7\% under black frames, question text only, and options only (Figure~\ref{fig:diagnostics}d). Under forced choice without abstention, question-only $\KA$ is 27.9\%, against 66.7\% with video. We report $\KA$ for this control: on blind inputs $\HM$ can rise when models abstain more even as answers worsen. No text-only, options-only, or black-frame run reaches the random-abstention reference ($\HM=26.8$). The blind-input controls demonstrate a substantial dependence on visual evidence for answering Known questions.

\begin{figure}[t]
  \centering
  \includegraphics[width=\linewidth]{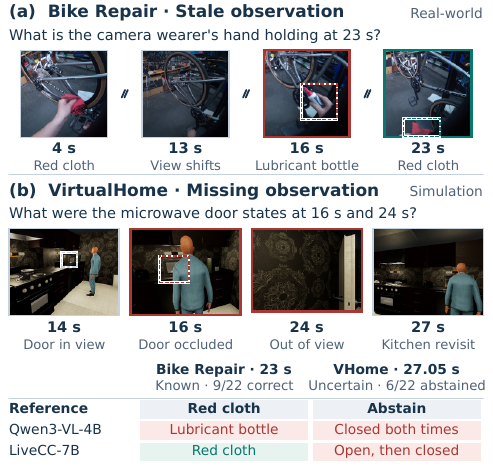}
  \caption{\textbf{Failure cases in real and simulated revisits.} (a)~The hand holds a lubricant bottle at 16\,s and a red cloth at 23\,s. Slashes mark cuts between visits. (b)~The kitchen revisit leaves the door states at 16 and 24\,s unresolved. Green and red cells mark correct and incorrect predictions. Full single-episode displays for all seven subsets appear in the supplementary material.}
  \label{fig:case-failures}
\end{figure}

\paragraph{Valid versus stale history.}
In the Bike Repair example, nine models select the lubricant bottle seen on an earlier visit despite the red cloth being visible at query time (Figure~\ref{fig:case-failures}a). To test how models use earlier evidence across revisits, we keep only the latest 16 frames. Truncation raises abstention and slightly hurts answering: median $\AR$ rises from 22.7\% to 34.6\%, median $\KA$ falls from 63.5\% to 60.9\%, and $\HM$ rises from 32.0 to 36.9. The two evidence windows respond in opposite ways (Table~\ref{tab:history-dependence}; per-model values in the supplementary material). Known items whose evidence starts earlier lose accuracy for all five models (median $\KA$ $67.8\rightarrow 57.2$), while items with evidence inside the window change little ($66.6\rightarrow 66.3$). Uncertain items gain abstention once the earlier views are removed, and the gain is larger when the uncertainty itself sits earlier ($22.5\rightarrow 40.2$) than when it sits in the last 16 frames ($21.9\rightarrow 25.8$). Earlier frames improve answering when historical evidence is needed, but retaining them also lowers abstention when the query remains unresolved after later revisits.

\paragraph{Frame order and future frames.}
Two further controls isolate the truncation effect. Shuffling frame order lowers median $\KA$ from 63.5\% to 57.4\% while leaving $\AR$ nearly unchanged: answering is more sensitive to temporal order than abstention under this control. As in the full-video check in Sec.~\ref{sec:rq2}, adding frames after $t_q$ changes $\KA$ and $\AR$ by at most 3.8 points per model in either window. Neither control matches the abstention gain from removing early frames.

\subsection{Discussion: Evidence Validity Across Revisits}
\label{sec:discussion}

\paragraph{Two failure modes under episodic revisits.}
Our results reveal two distinct challenges in using historical observations under dense revisits. The first is \emph{stale evidence}: an earlier observation was correct when recorded but is superseded by a later view. In Figure~\ref{fig:case-failures}a, the lubricant bottle observed at 16\,s becomes an incorrect answer once the hand is seen holding a red cloth at 23\,s. The second is \emph{insufficient evidence}: the observed history does not establish the requested fact because a relevant transition remains unseen. In Figure~\ref{fig:teaser}a, the same reappearance-order question changes from Uncertain at 41\,s to Known at 44\,s after an additional view of the empty plate location. These cases demand different behavior: updating an answer when new evidence is available, and abstaining when the evidence remains insufficient to settle the query.

\paragraph{Retaining history versus assessing its validity.}
Historical observations cannot simply be discarded, since earlier frames may contain evidence for Known questions. However, retaining more frames does not necessarily improve the model's ability to recognize when a revisit leaves the query unresolved. As shown in Table~\ref{tab:history-dependence}, restricting the input to the latest 16 frames reduces accuracy on questions requiring older evidence while increasing abstention on Uncertain questions. This contrast highlights a challenge of dense-revisit video understanding: useful evidence must be preserved, but its relevance must be assessed relative to the query time and subsequent observations. \bench{} evaluates these complementary capabilities through Known accuracy and abstention recall under a query-time protocol.

%% file: tables/tab_real_results.tex
\begin{table}[!t]
\centering
\caption{\textbf{Main evaluation on \bench{} (RQ1).} Per-subset and seven-subset avg.\ $\KA$/$\AR$/$\HM$ (\%). Ranked by avg.\ $\HM$ (bold/underline: best/second-best); $\HM$ is the unweighted mean of per-subset $\HM$. Random abstention is the chance baseline.}
\label{tab:real-results}
\scriptsize
\setlength{\tabcolsep}{1.2pt}
\renewcommand{\arraystretch}{1.04}
\providecommand{\best}[1]{\textbf{#1}}
\providecommand{\second}[1]{\underline{#1}}
\begin{tabularx}{\columnwidth}{@{}>{\raggedright\arraybackslash}X ccccccc cc >{\columncolor{stormhead!48}}c@{}}
\toprule
\rowcolor{stormhead!55}
& \multicolumn{7}{c}{\textbf{$\HM$ by subset}} & \multicolumn{3}{c@{}}{\cellcolor{stormhead}\textbf{7-subset avg.}} \\
\rowcolor{stormhead!55}
\textbf{Model} & Cook & Bike & Hlth & Mus. & Sprt & THOR & VHm. &
\cellcolor{stormhead}$\KA$ & \cellcolor{stormhead}$\AR$ & \cellcolor{stormhead!82!stormblue}$\HM$ \\
\midrule
Random abstention & 26.8 & 26.8 & 26.8 & 26.8 & 26.8 & 26.8 & 26.8 & 21.1 & 36.6 & 26.8 \\
\midrule
\rowcolor{stormhead!68}
\multicolumn{11}{@{\hspace{1.2pt}}l@{}}{\textbf{Offline video LLMs}\enspace\textcolor{stormgray}{\textit{Causal prefix re-encoded per query}}} \\
GLM-4.1V-9B & 52.4 & 45.3 & \second{59.3} & 42.7 & \second{54.6} & \best{48.8} & \best{51.9} & 42.2 & \second{72.1} & \best{50.7} \\
VideoLLaMA3-7B & \second{56.6} & \second{56.8} & 58.5 & 56.2 & \best{62.4} & 20.9 & 40.3 & 57.8 & 47.4 & \second{50.2} \\
Molmo2-8B & \best{59.1} & \best{59.5} & \best{65.2} & 53.6 & 50.0 & 20.0 & \second{43.7} & 51.0 & 60.3 & 50.1 \\
Qwen3.5-4B & 54.0 & 39.6 & 57.8 & \best{68.1} & 50.9 & 42.2 & 37.1 & 67.7 & 40.6 & 50.0 \\
Eagle2.5-8B & 40.6 & 41.4 & 48.9 & \second{60.9} & 50.7 & 25.4 & 41.3 & 62.4 & 36.7 & 44.2 \\
Qwen2.5-VL-7B & 51.3 & 50.8 & 54.1 & 30.7 & 50.6 & \second{46.3} & 15.9 & 31.2 & \best{80.0} & 42.8 \\
Qwen3-VL-32B & 46.5 & 43.9 & 55.3 & 59.7 & 50.9 & 11.1 & 26.1 & \second{67.7} & 32.1 & 41.9 \\
Qwen3-VL-4B & 40.0 & 34.1 & 50.0 & \best{68.1} & 42.8 & 17.3 & 13.9 & 67.1 & 28.5 & 38.0 \\
Qwen2.5-VL-3B & 43.9 & 39.7 & 42.2 & 34.2 & 38.2 & 38.4 & 16.5 & 32.5 & 47.5 & 36.1 \\
Qwen3.5-9B & 42.1 & 23.5 & 55.8 & 58.3 & 45.8 & 0.0 & 22.6 & \best{75.8} & 25.0 & 35.4 \\
MiniCPM-V~4.5 & 35.2 & 21.6 & 37.9 & 45.6 & 40.4 & 36.1 & 7.4 & 63.5 & 22.7 & 32.0 \\
InternVideo2.5-8B & 27.1 & 22.8 & 40.5 & 55.2 & 40.2 & 7.6 & 28.3 & 60.6 & 23.8 & 31.7 \\
Qwen3-VL-8B & 23.4 & 13.3 & 43.8 & 53.6 & 39.6 & 0.0 & 13.8 & 65.5 & 19.1 & 26.8 \\
InternVL3.5-8B & 16.6 & 6.0 & 27.4 & 27.5 & 17.1 & 0.0 & 7.4 & 66.0 & 8.7 & 14.6 \\
LLaVA-NeXT-V-7B & 6.6 & 2.0 & 4.3 & 7.3 & 14.3 & 0.0 & 0.0 & 38.4 & 2.9 & 4.9 \\
\rowcolor{stormhead!33}
\textit{Median} & 42.1 & 39.6 & 50.0 & 53.6 & 45.8 & 20.0 & 22.6 & 62.4 & 32.1 & \cellcolor{stormhead}38.0 \\
\midrule
\rowcolor{stormhead!68}
\multicolumn{11}{@{\hspace{1.2pt}}l@{}}{\textbf{Streaming models}\enspace\textcolor{stormgray}{\textit{Native state updated per frame}}} \\
Dispider-7B & 51.4 & 37.0 & 40.8 & 42.5 & 46.5 & 0.0 & 40.0 & 46.6 & 35.3 & 36.9 \\
LiveCC-7B & 47.7 & 47.0 & 44.0 & 36.0 & 43.0 & 7.5 & 23.4 & 36.8 & 41.7 & 35.5 \\
TimeChat-Online-7B & 43.7 & 29.3 & 44.6 & 51.6 & 46.8 & 3.9 & 23.3 & 55.0 & 28.6 & 34.7 \\
StreamChat-7B & 38.9 & 34.0 & 34.4 & 31.4 & 40.0 & 10.2 & 25.8 & 30.9 & 39.0 & 30.6 \\
Flash-VStream-7B & 38.8 & 30.2 & 24.8 & 32.6 & 39.9 & 26.1 & 17.8 & 45.6 & 23.7 & 30.0 \\
ReKV-7B & 32.7 & 25.4 & 23.7 & 30.1 & 44.5 & 7.6 & 13.4 & 56.1 & 17.6 & 25.3 \\
MovieChat-7B & 21.7 & 15.8 & 14.4 & 23.8 & 11.5 & 13.4 & 22.2 & 36.6 & 11.9 & 17.6 \\
\rowcolor{stormhead!33}
\textit{Median} & 38.9 & 30.2 & 34.4 & 32.6 & 43.0 & 7.6 & 23.3 & 45.6 & 28.6 & \cellcolor{stormhead}30.6 \\
\bottomrule
\end{tabularx}
\vspace{-1em}
\end{table}

%% file: tables/tab_abstention_mechanism.tex
\begin{table}[t]
\centering
\caption{\textbf{How models abstain (RQ3).} Seven-subset averages (\%). Cov./Risk/Prec.: coverage, selective risk, and the share of abstentions on Uncertain items; FAR: wrong abstention on Known; $\AR$ as in Sec.~\ref{sec:scoring}. $\KA$ with abstention ($+\bot$) vs.\ forced choice without it (Frc.). On Known items, abstention rate when forced choice would be wrong ($\bot|\times$) or right ($\bot|\checkmark$).}
\label{tab:abstention-mechanism}
\scriptsize
\setlength{\tabcolsep}{1.2pt}
\renewcommand{\arraystretch}{1.08}
\begin{tabularx}{\columnwidth}{@{}l
*{5}{>{\centering\arraybackslash}X}
*{2}{>{\centering\arraybackslash}X}
*{2}{>{\columncolor{stormhead!35}\centering\arraybackslash}X}@{}}
\toprule
\rowcolor{stormhead!55}
& \multicolumn{5}{c}{\textbf{Selective prediction}} & \multicolumn{2}{c}{\textbf{$\KA$}} &
\multicolumn{2}{c@{}}{\cellcolor{stormhead}\textbf{Abst.\ on Known}} \\
\rowcolor{stormhead!55}
\tabhead{Model} & Cov. & Risk & Prec. & FAR & $\AR$ & $+\bot$ & Frc. &
\cellcolor{stormhead}$\bot|\times$ & \cellcolor{stormhead}$\bot|\checkmark$ \\
\midrule
Q2.5-VL-3B & 63.7 & 61.4 & 33.5 & 32.0 & 47.5 & 32.5 & 45.2 & 32.1 & 26.8 \\
Q2.5-VL-7B & 39.9 & 44.0 & 35.6 & 53.0 & 80.0 & 31.2 & 53.2 & 60.0 & 36.9 \\
Q3-VL-8B & 93.1 & 48.0 & 67.0 & 2.0 & 19.1 & 65.5 & 67.6 & 2.2 & 0.9 \\
IVL3.5-8B & 96.4 & 49.5 & 59.2 & 1.5 & 8.7 & 66.0 & 66.7 & 1.7 & 0.9 \\
MCPM-V 4.5 & 89.0 & 47.5 & 57.9 & 6.5 & 22.7 & 63.5 & 67.2 & 9.0 & 3.6 \\
\midrule
\rowcolor{stormhead!33}
Median (5) & 89.0 & 48.0 & 57.9 & 6.5 & 22.7 & 63.5 & 66.7 & \cellcolor{stormhead!72}9.0 & \cellcolor{stormhead!72}3.6 \\
\rowcolor{stormhead!33}
Median (22) & 81.2 & 48.7 & 44.3 & 13.1 & 30.3 & 55.5 & -- & -- & -- \\
\bottomrule
\end{tabularx}
\end{table}

%% file: tables/tab_history_dependence.tex
\begin{table}[t]
\centering
\caption{\textbf{History dependence (RQ4).} Median $\KA$/$\AR$ of the five control models, pooled and split by evidence in the latest 16 frames before $t_q$ (Last 16) vs.\ earlier, plus seven-subset mean $\HM$ (\%) per input; parentheses give median paired per-model \mbox{$\Delta$ from the full prefix and item counts in the headers}.}
\label{tab:history-dependence}
\scriptsize
\setlength{\tabcolsep}{1.4pt}
\renewcommand{\arraystretch}{1.10}
\begin{tabularx}{\columnwidth}{@{}l
*{4}{>{\centering\arraybackslash}X}
>{\centering\arraybackslash}X@{}}
\toprule
\rowcolor{stormhead!55}
& \multicolumn{2}{c}{\textbf{$\KA$ (Known)}} & \multicolumn{2}{c}{\textbf{$\AR$ (Uncertain)}} & \\
\rowcolor{stormhead!55}
\tabhead{Input} & Last 16 & \cellcolor{stormhead}Earlier & Last 16 & \cellcolor{stormhead}Earlier & $\HM$ \\
\rowcolor{stormhead!55}
& {\tiny (1,163)} & \cellcolor{stormhead}{\tiny (739)} & {\tiny (329)} & \cellcolor{stormhead}{\tiny (346)} & \\
\midrule
Full prefix (default) & 66.6 & 67.8 & 21.9 & 22.5 & 32.0 \\
\rowcolor{stormhead!42}
Latest 16 frames & 66.3 {\tiny(-1.2)} & \cellcolor{stormhead}57.2 {\tiny(-9.1)} &
25.8 {\tiny(+2.1)} & \cellcolor{stormhead}40.2 {\tiny(+7.8)} & 36.9 {\tiny(+2.5)} \\
Shuffled frame order & 58.4 {\tiny(-5.8)} & 55.8 {\tiny(-10.0)} & 23.1 {\tiny(-0.3)} & 23.7 {\tiny(+1.2)} & 31.7 {\tiny(-1.1)} \\
Full video (with future) & 65.4 {\tiny(-1.5)} & 66.7 {\tiny(-0.9)} & 20.7 {\tiny(-1.2)} & 18.8 {\tiny(-0.9)} & 27.5 {\tiny(-1.7)} \\
\bottomrule
\end{tabularx}
\end{table}

%% file: contents/05_conclusion.tex
\section{Conclusion}
\label{sec:conclusion}

While video benchmarks have increasingly emphasized longer recordings, \bench{} shows that short videos remain challenging when scenes are revisited frequently. Across dense episodic revisits, earlier observations may become outdated, while important state changes can occur out of view. Our evaluation of 22 video LLMs reveals that current models still struggle with these challenges. Neither model scaling nor streaming memory consistently improves performance, and ranking by Known accuracy differs substantially from ranking by the combined score. Removing earlier frames further reveals that historical observations are valuable for answering questions but can also make models less likely to abstain when the available observations are insufficient. These findings highlight dense episodic revisits as an important challenge for video understanding and motivate models that can effectively track and update scene states across repeated visits. Higher-frame-rate sampling and open-ended answering are natural directions for future work.

%% file: contents/appendix_construction.tex
\section*{Contents of the Supplementary Material}
\label{sec:appendix-overview}

\newcommand{\apptocsec}[2]{%
  \par\vspace{0.45em}%
  \noindent
  {\color{stormblue}\rule[-0.12em]{2.2pt}{0.95em}}\hspace{0.45em}%
  \hyperref[#1]{\textcolor{stormdark}{\textbf{\ref*{#1}}}}%
  \hspace{0.55em}%
  \hyperref[#1]{\textcolor{stormdark}{\textbf{#2}}}%
  \ {\color{stormgray}\leaders\hbox to 0.52em{\hss.\hss}\hfill}\ %
  \hyperref[#1]{\textcolor{stormdark}{\textbf{\pageref*{#1}}}}%
  \par}
\newcommand{\apptocsub}[2]{%
  \par\vspace{0.12em}%
  \noindent\hspace{1.55em}%
  \makebox[2.15em][l]{\hyperref[#1]{\textcolor{stormgray}{\ref*{#1}}}}%
  \hyperref[#1]{#2}%
  \ {\color{stormgray}\leaders\hbox to 0.52em{\hss.\hss}\hfill}\ %
  \hyperref[#1]{\textcolor{stormgray}{\pageref*{#1}}}%
  \par}

{\small
\apptocsec{sec:appendix-data}{Dataset Construction and Statistics}
\apptocsub{sec:annotation}{Real-world episode construction}
\apptocsub{sec:appendix-sim-construction}{Simulated episode construction}
\apptocsub{sec:appendix-schema}{Annotation schema and field definitions}
\apptocsub{sec:appendix-qa-types}{Semantic question types}
\apptocsub{sec:appendix-cells}{Release statistics}
\apptocsub{sec:audit}{Data validation and quality audit}
\apptocsub{sec:appendix-license}{Licensing and release}
\apptocsec{sec:appendix-exp-details}{Experimental Details}
\apptocsub{sec:appendix-eval-protocol}{Evaluation protocol}
\apptocsub{sec:appendix-metric-formulas}{Metric formulas}
\apptocsub{sec:appendix-inference}{Models and inference settings}
\apptocsec{sec:appendix-extra-results}{Additional Experimental Results}
\apptocsub{sec:appendix-six-cells}{Complete six-cell decompositions}
\apptocsub{sec:appendix-sim-results}{Simulation results}
\apptocsub{sec:appendix-slice-analysis}{Global diagnostics}
\apptocsub{sec:appendix-input-controls}{Input controls}
\apptocsec{sec:appendix-cases}{Case Display}
\apptocsec{sec:llm-usage}{LLM Usage}
\apptocsec{sec:appendix-prompts}{Prompt Templates}
}

\section{Dataset Construction and Statistics}
\label{sec:appendix}
\label{sec:appendix-data}
\label{sec:appendix-construction}

\bench{} evaluates dense episodic revisits in short videos via two complementary tracks. The real-world track mines revisit episodes from HD-EPIC and Ego-Exo4D videos. The simulation track renders continuous camera rollouts with repeated observations and state changes in AI2-THOR and VirtualHome, supported by simulator records and observation-grounded annotations. Both tracks emit a 1~FPS stream, a query time $t_q$, a four-way question, evidence spans $W_i\subseteq[0,t_q]$, and diagnostic labels for accumulated change intensity and Known/Uncertain answerability.

\subsection{Real-world episode construction}
\label{sec:annotation}

\paragraph{Source domains.}
The real-world track combines Cook episodes from HD-EPIC~\citep{perrett2025hd} with Bike Repair, Health, Music, and Sports episodes from Ego-Exo4D~\citep{grauman2024egoexo4d}. Cook is the largest real-world domain (990 questions over 307 episodes, totaling 3.649 hours) and has the highest mean change intensity among the five (5.11; Table~\ref{tab:dataset-diagnostics}). Kitchen activities repeatedly relocate objects and transform ingredients, utensils, and appliances, producing dense state transitions. The remaining four domains add bike-repair, health, music, and sports contexts, and every domain is reported separately. Cook/Kitchen uses a four-stage pipeline (Figure~\ref{fig:real-pipeline}); prompts for both tracks appear in Appendix~\ref{sec:appendix-prompts}.
\input{contents/appendix_kitchen_pipeline}

\subsection{Simulated episode construction}
\label{sec:appendix-sim-construction}

\input{contents/appendix_sim_construction}

\subsection{Annotation schema and field definitions}
\label{sec:appendix-schema}

Every QA sample contains the core evaluation fields illustrated below: \texttt{answer\_index} stores the main answer, while \texttt{diagnostics} stores the Known/Uncertain label (\texttt{epistemic\_status}) and uncertainty sources. The item-level \texttt{change\_intensity} uses the track-specific rule defined below. The release stores one item per line (JSONL); the example below is a Bike release item with provenance omitted. Paired simulation examples appear in Appendix~\ref{app:sim-validation} and Appendix~\ref{app:sim-virtualhome}.

\begin{tcolorbox}[
  enhanced,
  breakable,
  colback=stormhead!55,
  colframe=stormblue!55!stormdark,
  boxrule=0.45pt,
  arc=1.2mm,
  left=1.4mm,
  right=1.4mm,
  top=0.8mm,
  bottom=0.8mm,
  before skip=0.5em,
  after skip=0.7em
]
\begin{Verbatim}[
  fontsize=\scriptsize,
  breaklines=true,
  breakanywhere=true,
  breaksymbolleft={},
  breaksymbolright={},
  formatcom=\color{stormdark}
]
{
  "id": "RevisitQA_bike_03_q02",
  "episode_id": "RevisitQA_bike_03",
  "query_time": 30.0,
  "question_type": "current_state",
  "question_subtype": "spatial_relation_at_query",
  "video_evidence": "At 29.0s both hands press the tan-sidewall tire onto the black rim; at 30.0s the assembled wheel is held up showing the tire around the rim.",
  "question": "At 30.0s, how is the tan-sidewall tire positioned relative to the black rim while held in both hands?",
  "options": [
    "The tire is held beside the rim with no portion overlapping the rim",
    "The tire encircles the rim with both sidewalls over the rim edges",
    "The tire is folded against one side of the rim with only a sidewall touching it",
    "Cannot be determined from the video up to this point."
  ],
  "answer_index": 1,
  "abstention_index": 3,
  "evidence_spans": [[29.0, 30.0]],
  "diagnostics": {
    "epistemic_status": "known",
    "uncertainty_sources": []
  },
  "diagnostic_rationale": {
    "volatility": "After being separated, the tire is worked back onto the rim, so the state changes over the episode.",
    "uncertainty": "The later frames clearly show the tire seated around the rim before inflation."
  },
  "change_intensity": 4
}
\end{Verbatim}
\end{tcolorbox}

\label{sec:appendix-fields}
For real-world episodes, \texttt{change\_intensity} is the number of selected visit intervals whose output start time is no later than \texttt{query\_time}, with a minimum value of one. For THOR, it counts event intervals from their before-observation boundary; for VHome, it counts the recorded event starts no later than the query. Both use a minimum of one (Eq.~\ref{eq:sim-intensity}). These two track-specific ordinal proxies share the common range 1--10 and bins 1--3 ($B_1$), 4--6 ($B_2$), and 7--10 ($B_3$).

\texttt{epistemic\_status} is \texttt{known} (Known) when $V_{\le t_q}$ supports one substantive answer and \texttt{uncertain} (Uncertain) when abstention is correct. For Uncertain items, \texttt{uncertainty\_sources} records why the prefix is insufficient: missing or partial observation, low visual quality, ambiguous evidence or attribute, or \mbox{multiple candidates}.

\subsection{Semantic question types}
\label{sec:appendix-qa-types}

Table~\ref{tab:qa-types} summarizes the six semantic question types in the real-world track. THOR and VHome use the same type names with track-specific builders (Table~\ref{tab:sim-qa-rules}; Appendices~\ref{app:sim-thor-paired} and~\ref{app:sim-virtualhome}): THOR room identification, reappearance-order pairs, and category returns; VHome visibility, return/exit counts, and reviewed physical-state items. Each item has one semantic type; change-intensity bin and Known/Uncertain status are annotated independently.
\input{tables/tab_qa_types}

\subsection{Release statistics}
\label{sec:appendix-cells}

Table~\ref{tab:main-stats} lists per-subset source, scale, hours, duration, QA density, Known/Uncertain and change-intensity counts, and the Earlier evidence-window share. Table~\ref{tab:dataset-diagnostics} expands the release with question types, uncertainty sources, mean change intensity, query time, and evidence-span diagnostics.

Stratified metrics are computed over all populated (change-intensity $\times$ answerability) cells within each domain.

\input{tables/tab_main_stats}
\input{tables/tab_dataset_diagnostics}

\subsubsection{Duration and QA density}
\label{sec:appendix-density}

Figure~\ref{fig:teaser}b compares median durations and QA density; Table~\ref{tab:overview-density} lists the underlying statistics. QA density divides the supplied QA counts by total video time.

\input{tables/tab_overview_density}

\subsection{Data validation and quality audit}
\label{sec:audit}

Audits start from the generated QA pool and check that answers depend on the video, that labels follow $V_{\le t_q}$, and that evaluation exposes only frames up to $t_q$. Removed and relabeled item IDs are released with the benchmark.

\paragraph{Release composition and integrity.}
Consecutive revisit windows from the same long video are temporally thinned by retaining alternating windows (\texttt{001}, \texttt{003}, \ldots), and the associated video and QA records are removed jointly. QA generation targets roughly four Known items per Uncertain item within each question type; after the audits below, the release contains 2,577 items from 622 episodes, 1,902 Known and 675 Uncertain (26.2\%, Table~\ref{tab:main-stats}). A release-level integrity check covers schema conformance, globally unique identifiers, three distinct substantive options plus the shared abstention option, a valid answer index, legal question types and uncertainty sources, referenced-video availability, and integer change-intensity labels in $[1,10]$. Every Uncertain item has a nonempty uncertainty-source set and uses the abstention option as its answer. Every evidence span satisfies $t_{\mathrm{start}}\leq t_{\mathrm{end}}\leq t_q$, and the evaluation loader independently restricts frame access to timestamps at or before $t_q$. Evaluation applies a deterministic, sample-specific permutation of the three substantive options, identical for all models, while the abstention option keeps its position; CIs use 1,000 episode-clustered bootstraps.

\subsubsection{Non-visual shortcuts}

\paragraph{Option structure.}
Every item has three substantive options and the same abstention option (\emph{Cannot be determined from the video up to this point}) in a fixed last position.

\paragraph{Surface cues in the text.}
Within each domain, question type, and Known/Uncertain class, we compare option lengths, unique longest/shortest rates, and question--option lexical overlap. Flagged items are revised for comparable length, syntax, and specificity; distractors stay in the same semantic frame and differ in a visually grounded property. Stems that imply the answer, abstention-like distractors, and spelled-out counts are rewritten. In the release, the rate at which the correct option is uniquely longest stays near chance (33.3\%) across subsets.

\paragraph{Text predictability filtering.}
A bag-of-words classifier predicts Known versus Uncertain from the question and substantive options under episode-grouped five-fold cross-validation. In the real-world track, its out-of-fold scores drive a constrained filter that removes the items most predictable from text while holding the Known share within 1.5 points and the question-type, change-intensity, evidence-age, and uncertainty-source shares within 2 points of their pre-filter values; no episode is lost.

\paragraph{Model text priors.}
Five video LLMs (Qwen2.5-VL-3B/7B, Qwen3-VL-8B, InternVL3.5-8B, MiniCPM-V-4.5) answer every Known item without video or abstention. Items solved by $\ge$4 of 5 are removed (chance 4.5\%). Restoring the removed items raises filter-model video $\KA$ by 6.4 points.

\paragraph{Paired and time-balanced simulation items.}
In simulation, Known and Uncertain items share question and option sets wherever possible, so the label depends only on the prefix. For THOR reappearance-order items, each 2\,s query-time bin is balanced (query-time AUROC~$0.50$).

\paragraph{Input controls.}
Options-only, text-only, black-frame, shuffled-frame, and forced-choice inputs measure what each model can do without the video (Section~\ref{sec:rq4}, Table~\ref{tab:input-controls}).

\subsubsection{Label verification}

\paragraph{Labeling rubric.}
All label checks follow one rubric. A fact is Known when exactly one candidate is supported by frames up to $t_q$ and the other two are ruled out. Observed before and after states establish a change even if the action itself is off-screen. Questions restricted to what is visible are answered from the frames. Questions about what actually happened become Uncertain when an unobserved gap could hide a relevant event. A current-state question requires the state to be observed after the last moment it could have changed. Identity across a gap requires a distinguishing mark, continuous tracking, or the absence of look-alikes. Text, digits, and small objects count as visible only if legible at the evaluation resolution ($\approx$448$\times$448). Items that no video could answer as posed are marked invalid. Uncertain items record one or more of the six uncertainty sources.

\paragraph{Simulation labels.}
Simulation labels are computed from execution traces and rendered masks and then checked on the frames. A reappearance-order question is Known only when one object is seen back at its place and a later frame up to $t_q$ still shows the other place empty; otherwise the other object may have returned while the camera was away. VHome count questions are Known only when every counted action is fully\mbox{ observed}.

\paragraph{Cross-model verification.}
After the shortcut filters above, the remaining candidates enter blind label verification. Let the pool label denote the construction-time Known/Uncertain answerability label. Each item is packaged without that label as a frame package truncated at $t_q$ (1\,FPS timeline with timestamps, original-resolution frames at $t_q$ and the three preceding seconds, the $t_q$ frame at evaluation resolution, and the question with its three substantive options). GPT-6.1 Sol blindly labels the filtered pool (2,655 real-world, 278 VHome, and 196 THOR reappearance-order candidates). Claude Opus 5.5 independently labels all disagreements with the pool label, plus a 15\% sample of agreeing items. Final labels follow a majority vote over the pool label and available model judgments; unresolved items are discarded, leaving 2,436. The remaining 141 THOR Known builders outside the reappearance-order family are labeled from simulator traces and masks, then pass the text-prior filter and per-bin balance, and join to give the 2,577-item release.

\paragraph{Final audit.}
A stratified audit of 141 items by two independent annotators yielded 84.4\% inter-annotator agreement (Cohen's $\kappa=0.688$). Agreement with the release labels was 85.1\% and 83.7\%, respectively.

\subsection{Licensing and release}
\label{sec:appendix-license}

The real-world episodes are derived from HD-EPIC~\citep{perrett2025hd} and Ego-Exo4D~\citep{grauman2024egoexo4d} and remain subject to their licenses; access to the derived episodes follows the same terms. The simulated episodes are rendered with AI2-THOR~\citep{kolve2017ai2thor} and VirtualHome~\citep{puig2018virtualhome}. Annotations, the removed and relabeled item IDs from the audits, and the evaluation code are released with the benchmark.

\FloatBarrier
\section{Experimental Details}
\label{sec:appendix-exp-details}
\label{sec:metric-definitions}

\subsection{Evaluation protocol}
\label{sec:appendix-eval-protocol}

Let $V=(v_1,\dots,v_T)$ be the 1~FPS stream and $t_q$ the query time. Under the causal-prefix protocol, models see only $V_{\le t_q}$ (Section~\ref{sec:protocol}). The full-video offline control provides the complete stream, including frames with $t > t_q$. Sample-specific deterministic permutations of the substantive options remove positional priors. Each item is decoded independently in one call over the four options, including the shared abstention option.

\subsection{Metric formulas}
\label{sec:appendix-metric-formulas}

$\KA$, $\AR$, and $\HM$ follow the main paper; per-cell values restrict the sums to items in change-intensity bin $b$ with state $s$. We additionally report false-abstention rate on Known items $\mathrm{FAR}=|\mathcal{K}|^{-1}\sum_{i\in\mathcal{K}}\mathbb{1}[\hat{a}_i=\bot]$, coverage $\mathrm{Cov}=N^{-1}\sum_{i=1}^{N}\mathbb{1}[\hat{a}_i\neq\bot]$, and risk among answered items $\mathrm{Risk}=\bigl(\sum_{i}\mathbb{1}[\hat{a}_i\neq\bot\land\hat{a}_i\neq a_i]\bigr)/\bigl(\sum_{i}\mathbb{1}[\hat{a}_i\neq\bot]\bigr)$.

\paragraph{Chance level of $\HM$.}
A policy that abstains with probability $p$ and otherwise guesses uniformly among the three substantive options has $\KA=(1-p)/3$ and $\AR=p$. $\HM$ is maximized at $p\approx0.366$ ($\KA=21.1$, $\AR=36.6$, $\HM=26.8$; Table~\ref{tab:real-results}). Always abstaining or never abstaining yields $\HM=0$. Effective reliability~\citep{whitehead2022reliable} scores correct/wrong/abstain as $+1/-1/0$ and gives correct Uncertain abstention no credit; $\HM$ keeps both axes and stays zero for the two trivial policies.

\paragraph{Ranking agreement across metrics.}
On seven-subset averages over 22 models, Kendall's $\tau$ is 0.06 between $\KA$ and $\HM$ and 0.73 between $\AR$ and $\HM$, so $\HM$ rankings are not interchangeable with answer accuracy alone. Question-count weighting of the seven subsets agrees with equal weighting at $\tau=0.85$, with the same top four models by $\HM$.

\subsection{Models and inference settings}
\label{sec:appendix-inference}

We evaluate the 22 video LLMs in Section~\ref{sec:setup} (15 offline and 7 streaming) with zero-shot prompting. A standardized 1~FPS temporal clock supplies frames across all checkpoints; Table~\ref{tab:frame-budget} lists individual frame budgets and temporal units. Streaming models run in their native mode through one interface: \texttt{reset} at an episode boundary, \texttt{observe} with the frames that became visible since the previous query, and \texttt{answer} from the current state. Each frame is processed once; the question pass does not modify the state, and the answer is the option letter with the highest score. Weights are checked against released checkpoints after loading. Table~\ref{tab:streaming-models} lists each model's state mechanism and 1\,FPS clock-alignment settings under the shared $t_q$ protocol.

\input{tables/tab_frame_budget}
\input{tables/tab_streaming_models}

Decoding is greedy ($\text{do\_sample}=\text{False}$, $\text{num\_beams}=1$). Non-thinking adapters use an 8-token generation limit. GLM uses a 512-token thinking budget, followed by up to 64 tokens for answer completion when needed. Maximum frame resolution is capped at 200,704 pixels ($448\times 448$, $\approx$256 visual tokens per frame). Context window overflows apply a uniform keep-last frame truncation policy. Confidence intervals are the 2.5th--97.5th percentiles of 1,000 episode-clustered bootstraps.

Each item is one model invocation. The prompt presents the question stem with four options (A--D), where the last is the shared abstention option, and instructs the model to choose the option stating that the answer cannot be determined when the video so far does not provide enough evidence. Option parsing extracts a letter with regular expressions; an unparseable response counts as a wrong answer. Parsed responses are valid in 99.97\% of model--subset--protocol runs on average. The options-only control removes the question stem; the text-only control removes the video; the black-frame control replaces every visible frame with a black frame of the same size and timestamp. The forced-choice controls remove the abstention option and instruct the model to pick one of the three substantive options, with and without video. In the full-video offline control, models receive $V_{\le T}$ with the same prompt text and without a query-boundary marker.

\section{Additional Experimental Results}
\label{sec:appendix-extra-results}

\subsection{Complete six-cell decompositions}
\label{sec:appendix-six-cells}
\label{sec:appendix-cook-cells}

Table~\ref{tab:rq2-offline-br} details online versus full-video offline performance. Table~\ref{tab:diagnostic-cells} gives the Cook six-cell $\KA$/$\AR$ decomposition ($n\ge 29$ per cell); Table~\ref{tab:global-diagnostics} gives seven-subset diagnostics. Per-cell scores use episode-clustered bootstrap confidence intervals. Figure~\ref{fig:diagnostics}a shows $\KA$ and $\AR$ across change-intensity bins.
\input{tables/tab_rq2_offline_br}
\input{tables/tab_diagnostic_cells}

\subsection{Simulation results}
\label{sec:appendix-sim-results}

Question-count weighting of the seven subsets agrees with equal weighting at $\tau=0.85$ and keeps the same top four by $\HM$ (Appendix~\ref{sec:appendix-metric-formulas}). Table~\ref{tab:sim-results} provides the 22-model simulation evaluation matrix.
\input{tables/tab_sim_results}

\subsection{Global diagnostics}
\label{sec:appendix-slice-analysis}

Table~\ref{tab:global-diagnostics} reports seven-subset diagnostics, and Figure~\ref{fig:diagnostics}c summarizes question types. Table~\ref{tab:rq3-sources} details median performance and support counts across the six uncertainty root causes.
\input{tables/tab_global_diagnostics}
\input{tables/tab_rq3_sources}

\subsection{Input controls}
\label{sec:appendix-input-controls}
\label{sec:appendix-cook-ablations}

Figure~\ref{fig:diagnostics}d summarizes the input controls over the seven subsets; Table~\ref{tab:input-controls} reports per-model $\KA$, $\AR$, and $\HM$. We report $\KA$ and $\AR$ alongside $\HM$ to distinguish answer accuracy from abstention behavior. Table~\ref{tab:stale-history} gives the per-model values behind the evidence-window split in Table~\ref{tab:history-dependence}.
\input{tables/tab_input_controls}
\input{tables/tab_stale_history}

\FloatBarrier
\section{Case Display}
\label{sec:appendix-cases}

Figures~\ref{fig:case-cook}--\ref{fig:case-vh} show one released episode per subset, with two query times in THOR. The Bike Repair and VHome items also appear in condensed form in Figure~\ref{fig:case-failures} of the main paper. Each timeline spans the episode: black ticks mark cuts between real-world visits, and a gray segment marks frames after the final query. In THOR, the green interval from 41 to 44\,s adds the 43\,s view to the prefix. Green frame borders mark visual evidence; red borders mark stale views or observation gaps. Qwen3-VL-4B receives the sampled prefix at each query, and LiveCC-7B receives the frames in sequence. Choice letters follow the displayed option order; unsupported marks substantive answers on Uncertain items.

\begingroup
\setlength{\abovecaptionskip}{1pt}
\setlength{\belowcaptionskip}{0pt}
\newcommand{\stormcase}[3]{%
  \begin{center}%
    \includegraphics[width=\linewidth]{#1}%
    \captionof{figure}{#2}%
    \label{#3}\vspace{0.5em}%
  \end{center}%
}
\stormcase{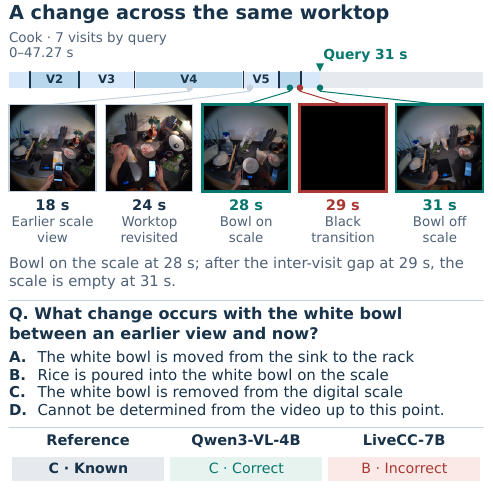}{\textbf{Cook.} Bowl on the scale at 28\,s; after the inter-visit gap at 29\,s, the scale is empty at 31\,s.}{fig:case-cook}
\stormcase{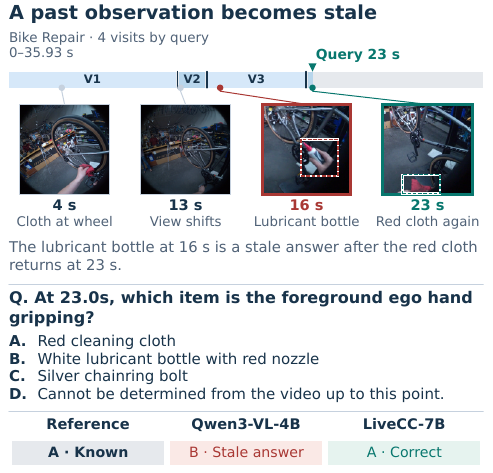}{\textbf{Bike Repair.} Lubricant at 16\,s vs.\ red cloth at the 23\,s query: earlier bottle is stale.}{fig:case-bike}
\stormcase{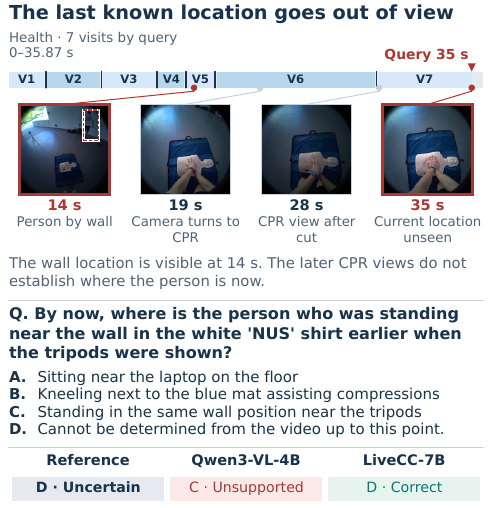}{\textbf{Health.} Person by the wall at 14\,s; later CPR views leave that location unknown.}{fig:case-health}
\stormcase{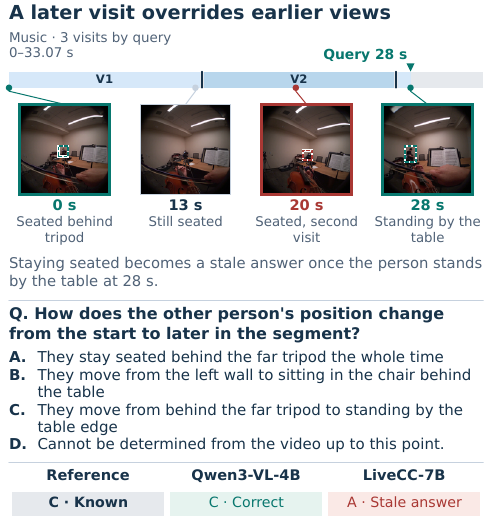}{\textbf{Music.} Seated behind the tripod through 20\,s, then standing by the table at 28\,s.}{fig:case-music}
\stormcase{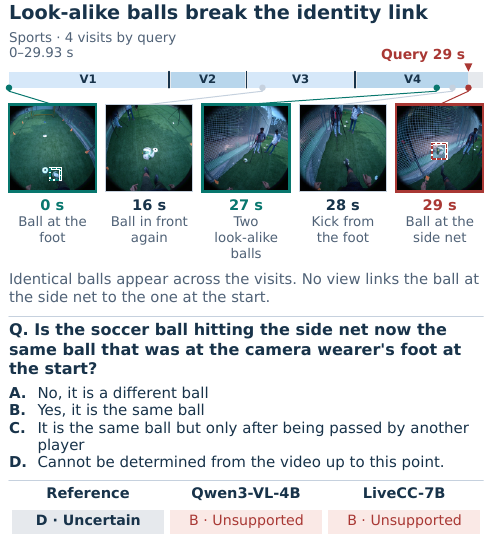}{\textbf{Sports.} Look-alike balls across four visits; no link from side-net ball at 29\,s to the earlier foot ball.}{fig:case-sports}
\stormcase{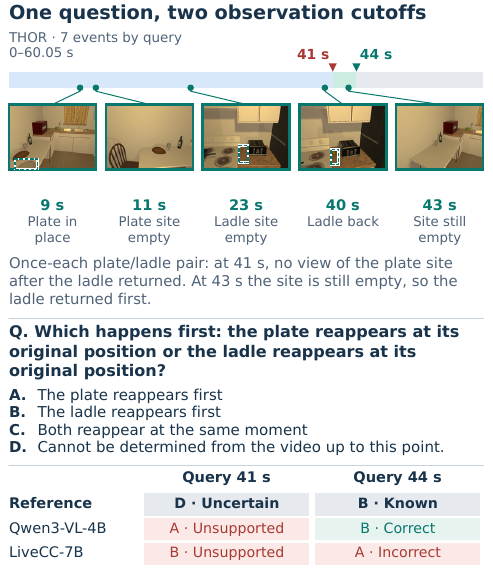}{\textbf{THOR.} Shared question/options: Uncertain at 41\,s, Known at 44\,s once the plate site is empty at 43\,s.}{fig:case-sim}
\begin{center}\vspace{-0.55em}
  \includegraphics[width=0.92\linewidth]{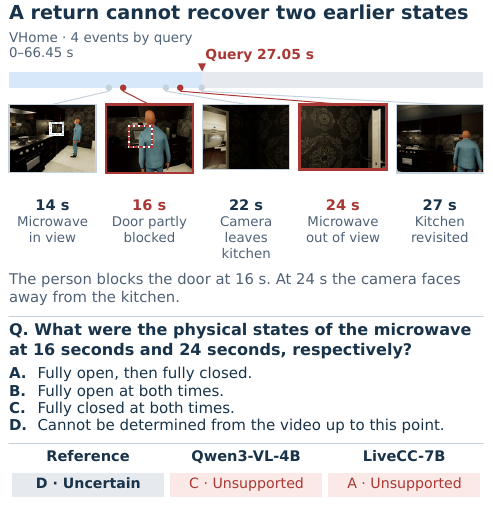}
  \captionof{figure}{\textbf{VHome.} Door blocked at 16\,s and camera away at 24\,s; both states stay unresolved on revisit.}
  \label{fig:case-vh}
\end{center}
\endgroup

\section{LLM Usage}
\label{sec:llm-usage}

Consistent with the main paper, the concrete uses are:
\begin{itemize}[leftmargin=*, itemsep=1pt, topsep=2pt]
\item \textbf{Real-world QA drafting.} Doubao Seed 2.1 Pro proposed candidate questions; multimodal LLMs proposed distractors (Appendix~\ref{sec:appendix-prompts}).
\item \textbf{Simulation.} VLMs refined template wording during development (Appendix~\ref{app:sim-template-refinement}); release items come from deterministic builders and simulator logs (Appendix~\ref{app:sim-validation}).
\item \textbf{Text edits.} Claude Opus 5.5 rewrote stems and distractors flagged by the text audits without changing the grounded fact, temporal scope, or label.
\item \textbf{Label checks.} GPT-6.1 Sol and Claude Opus 5.5 performed conditional label cross-checks; two human annotators audited a stratified sample under one rubric on packages truncated at $t_q$ (Appendix~\ref{sec:audit}).
\item \textbf{Text-prior filter.} Five open video LLMs answered Known items without video.
\item \textbf{Writing.} LLMs assisted proofreading and \LaTeX{} macros. Metrics use deterministic scripts on raw predictions.
\end{itemize}

\input{contents/appendix_prompts}

%% file: contents/appendix_kitchen_pipeline.tex
\paragraph{Release format of generated items.}
\label{app:kitchen-release-format}
QA generation produces four candidate options per item; for Uncertain items, the answer states why the fact cannot be determined. A formatting step then converts each item into the release format. On Uncertain items, the shared abstention option, \emph{``Cannot be determined from the video up to this\mbox{ point''}}, replaces this item-specific answer. On Known items, one distractor is dropped, abstention-like distractors first. Every released item therefore has three substantive options followed by the shared abstention option in a fixed last position. Text edits, filtering, and audits are in Appendix~\ref{sec:audit}.

\subsubsection{Construction Pipeline}
\label{app:kitchen_pipeline}

\begin{figure*}[t]
  \centering
  \includegraphics[width=\textwidth]{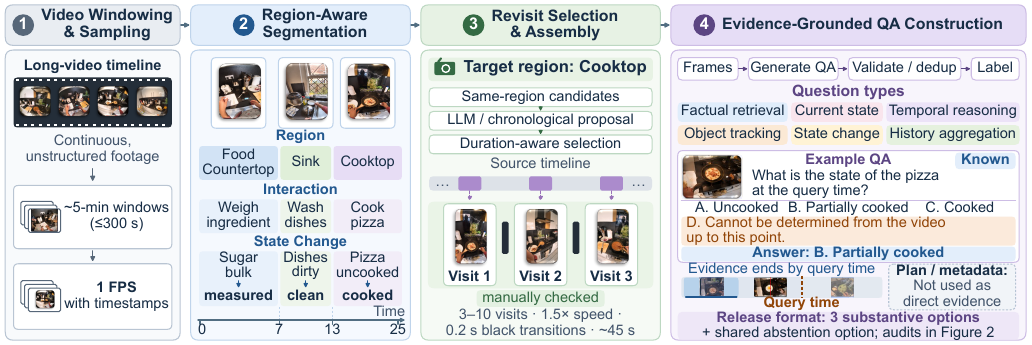}
  \caption{\textbf{Real-scene episode construction pipeline.} (1)~Source videos split into $\le$300\,s windows at 1~FPS. (2)~A VLM builds a region-aware timeline of regions, interactions, and state changes. (3)~Same-region visits are selected, checked, and assembled into $\approx$45\,s episodes. (4)~QA is generated from timestamped frames and labeled with evidence ending by $t_q$.}
  \label{fig:real-pipeline}
\end{figure*}

Figure~\ref{fig:real-pipeline} illustrates the four stages described below.

\paragraph{Windowing and semantic segmentation.}
Each selected HD-EPIC kitchen video is divided into consecutive processing windows of at most 300 seconds. A final remainder of at least 45 seconds forms an independent window, whereas a shorter remainder is merged into the preceding window. Frames are uniformly sampled at 1~FPS using absolute source-video timestamps. For semantic segmentation, timestamped frames are submitted to Doubao Seed 2.1 Pro in chronological batches. The predicted intervals jointly form a gap-free, region-aware timeline for the complete processing window; each interval records one functional region, one segment role, one activity type, zero or more visible change types, and a concise description.

\paragraph{Region-scoped revisit selection.}
The timeline is grouped by functional region. We discard unusable intervals as well as intervals assigned to \texttt{other\_area}. An interval is an eligible visit when it is labeled as either observation or interaction, belongs to a named kitchen region, and lasts at least 3 seconds. If a region is not specified manually, the pipeline selects one according to its usable duration, temporally separated occurrences, and number of eligible candidates. Segment boundaries and the selected visits were checked manually before assembly. Because visits are non-overlapping and kept in source order, each transition between two visits stands for real elapsed time in the source video, during which the region may change off camera. For the selected region, the pipeline produces both an LLM-selected proposal and a deterministic chronological proposal. Selection prefers visits that expose state differences across selected revisits and retains the proposal whose estimated duration is closest to 45 seconds when both are valid.

\paragraph{Video assembly.}
The retained visits are kept in source order and assembled into a short revisit episode with 0.2\,s black transition markers for omitted off-camera intervals, then standardized to $\sim$45\,s (1.5$\times$ playback) before 1~FPS evaluation sampling. The episode preserves the source resolution, removes audio, and is encoded with H.264. A companion plan JSON stores the region, I/O intervals, proposal source, and gap notes.

\paragraph{QA generation and item annotation.}
QA generation takes the assembled episode as visual input. The episode is sampled at 1~FPS; each frame is paired with its timestamp and supplied chronologically to a multimodal model. The model generates four-choice items spanning factual retrieval, current state, state change, object tracking, temporal reasoning, and history aggregation. Each item stores query time, type/subtype, choices, answer index, visible-evidence text, evidence spans, status, uncertainty sources, and rationales. These generated items are converted to the release format in Appendix~\ref{app:kitchen-release-format} before auditing. The item-level \texttt{change\_intensity} is derived from the revisit plan and follows the track-specific definition in Appendix~\ref{sec:appendix-fields}. Items are stored in per-episode JSON files and aggregated into a release-level JSONL manifest.

%% file: contents/appendix_sim_construction.tex
\begin{figure*}[t]
  \centering
  \includegraphics[width=\textwidth]{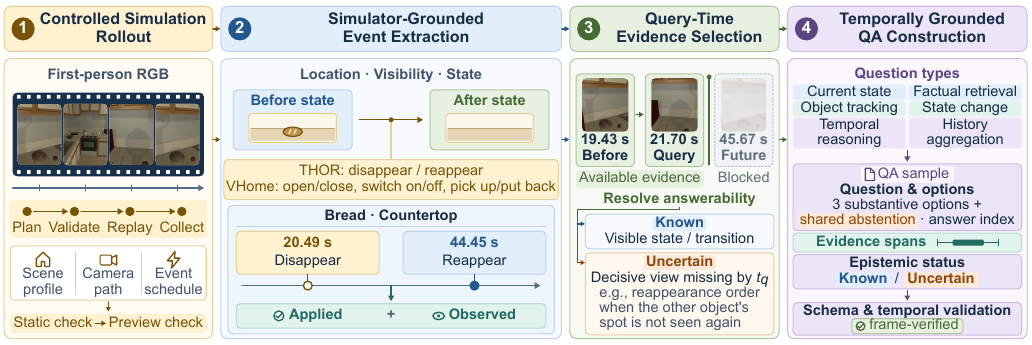}
  \caption{\textbf{Simulated-scene construction pipeline.} (1)~Planned and validated rollouts are replayed to record first-person RGB. (2)~Simulator states yield location, visibility, and state events that are both applied and observed. (3)~Evidence up to $t_q$ decides Known or Uncertain; later frames are blocked. (4)~Template-built items carry options, evidence spans, and Known/Uncertain labels and pass schema and temporal checks. Panels illustrate THOR; VHome follows the same stages with its own environment-specific event types (Appendix~\ref{app:sim-virtualhome}).}
  \label{fig:sim-pipeline}
\end{figure*}

The simulation track provides continuous simulated camera rollouts with two subsets: THOR (309 questions over 28 AI2-THOR episodes) and VHome (210 questions over 29 VirtualHome episodes). THOR covers the four room categories with seven episodes each, contributing 90, 73, 74, and 72 questions in Kitchen, Living Room, Bedroom, and Bathroom. Figure~\ref{fig:sim-pipeline} details the construction summarized in the simulation lane of Figure~\ref{fig:framework}. Seeded scene exploration and rendering produce event records. Fixed templates are polished with VLM assistance during development, then instantiated by deterministic builders into the shared annotation schema.

The following rollout and template details describe THOR; Appendix~\ref{app:sim-virtualhome} gives the VHome construction.

\subsubsection{Scene selection and rollout configuration}
\label{app:sim-rollout}

\paragraph{Scene and target selection.}
The scene profiler retrieves reachable camera positions and candidate pickupable objects on room-specific supporting surfaces. These include counters, dining tables, and stove burners in kitchens; tables, sofas, and TV stands in living rooms; beds, desks, dressers, and shelves in bedrooms; and sinks, bathtubs, shelves, and towel holders in bathrooms. Candidate viewpoints are rendered to check target visibility and to find an alternative orientation that excludes the target. Simulator-verified paths connect these viewpoints into a route; Table~\ref{tab:sim-render-settings} lists scenes and room parameters for all four runs.

\begin{table*}[t]
\centering
\caption{\textbf{Simulation rollout settings.} THOR scene numbers are \texttt{FloorPlan} suffixes (seven videos and 72--90 QA items per room category). Visibility thresholds apply when the queried target must stay in frame. VHome uses native VirtualHome capture.}
\label{tab:sim-render-settings}
\scriptsize
\setlength{\tabcolsep}{3.5pt}
\renewcommand{\arraystretch}{1.06}
\begin{tabularx}{\textwidth}{@{}ll*{4}{>{\centering\arraybackslash}X}@{}}
\toprule
\rowcolor{stormhead!55}
Room & Scene IDs & Dist.\ (m) & Speed & Step & Area/box \\
\midrule
Kitchen & 7, 12, 13, 19, 20, 27, 28 & 0.5--1.2 & 0.85 & 3.5 & 1400/28 \\
Living & 202, 209, 211, 221--223, 229 & 0.7--2.2 & 2.20 & 8.0 & 600/18 \\
Bedroom & 301, 306, 308, 318, 327, 329, 330 & 0.6--2.0 & 1.60 & 5.0 & 600/18 \\
Bathroom & 401, 404, 406, 408, 410, 422, 428 & 0.5--1.2 & 1.30 & 4.0 & 500/16 \\
\bottomrule
\end{tabularx}
\vspace{0.35em}

\noindent{\scriptsize\textit{VHome native capture:} 7 scenes / 29 rooms (7/7/8/7); $640\times480$, 20\,FPS (mean 66.91\,s); 10 events, 3--5 targets; 1 observer + 1--2 actors; Open/Close, SwitchOn/Off, Grab/PutBack ($\geq$4/room); QA visibility at 1\,FPS with $\geq128$\,px visible, $0$ absent, 2-sample transition confirm.}
\end{table*}

\paragraph{Video and motion parameters.}
Each episode follows a 60-second plan at $640\times480$ resolution, 20~FPS, and a $55^{\circ}$ field of view. Retaining both endpoints produces 1,201 frames and an encoded duration of 60.05 seconds. The camera moves between observation stations while turning toward and away from the selected objects. Common constraints require at least 8\,m of travel, a stationary fraction at most 0.18, a maximum yaw rate of $210^{\circ}$/s, and a maximum yaw acceleration of $1{,}000^{\circ}$/s$^2$. Room-specific speed and visibility thresholds fit layout and object-size differences.

\paragraph{Controlled changes.}
The released episodes instantiate ten events each: five disappearances and five reappearances, for 280 events overall. A disappearance removes a target from the scene through the simulator API \texttt{DisableObject}; a reappearance restores it through \texttt{EnableObject}. Both interventions are applied while the target is outside the camera view. The resulting sequence exposes a visible state before the intervention and a visible state after the camera returns, while the intervening operation remains unobserved. At trigger time $T$, the plan schedules a before observation at $T-1.05$\,s, an out-of-view check at $T-0.10$\,s, and an after observation at $T+1.20$\,s. The first trigger must occur by 4\,s, the last at or after 50\,s, and successive triggers must be no more than 8\,s apart.

\subsubsection{Plan validation and event extraction}
\label{app:sim-events}

\paragraph{Plan, validate, replay, and collect.}
The generator compiles scene selection, event times, camera orientations, and movement into a fixed trajectory. Static checks validate timing, frame count, travel, speed, and camera-motion constraints. A rendered preview checks the planned event views before full replay. The final replay writes the RGB video, execution trace, and before/after instance-mask evidence. An event is accepted when its action was applied and its visual transition was observed. Disappearance requires a qualifying visible target before the change and absence afterward; reappearance requires the reverse. The out-of-view check requires zero target-mask pixels. The episode is retained only when all event and trajectory checks pass. Failed attempts are retried with another deterministic seed. Each room batch requires distinct scenes, seeds, and event programs; all retained batches pass their recorded checks.

\paragraph{Event representation.}
For each accepted event, the exporter joins the event plan, execution record, and scene profile. It stores the object category, supporting surface, event type, before/after state descriptions, and the interval between the planned observation times. Internal target identifiers pair disappearances with reappearances. These identifiers support construction and provenance and only select objects that disappear and reappear exactly once; model-facing questions ask about visible categories, locations, or relations rather than physical identity. The event exporter instantiates descriptions from the accepted appearance changes.

\subsubsection{Query-time evidence and question construction}
\label{app:sim-qa}

\paragraph{Sampling and temporal boundaries.}
The QA exporter maps event before/after times to preceding/following integer samples at 1~FPS on $\{0,\dots,60\}$ within the camera observation dwell intervals. This aligns evidence boundaries with the exact 1~FPS sampled frames exposed to models during evaluation, where before/after target visibility and absence are validated against the rendered instance masks. Visible-transition questions use the sampled after time as $t_q$. Return questions span disappearance or earlier return through the selected reappearance. Query-time paired items place $t_q$ after both disappearances (Appendix~\ref{app:sim-thor-paired}). Room-identification questions query final scene context at $t_q=60$\,s over the $[0,60]$\,s episode. Stored intervals and evaluation samples both end at or before~$t_q$.

\paragraph{Six semantic families.}
Table~\ref{tab:sim-qa-rules} specifies the question builders in both simulation subsets. THOR Uncertain items use query-time paired reappearance-order questions under one shared stem; VHome Uncertain items cover physical-state, count, and tracking families under occlusion. In THOR, Known current-state and factual-retrieval builders query scene context along the trajectory; the other Known builders use three distinct object categories from the same episode. History aggregation selects the last return in a local pair, while temporal reasoning asks for the second return in that pair; these two builders share event-order evidence. All 49 THOR Uncertain items use the query-time paired reappearance-order template below.

\begin{table*}[t]
\centering
\caption{\textbf{Simulation question construction and release counts.} K/U = Known/Uncertain. Known targets list THOR then VHome. All THOR Uncertain items use the paired reappearance-order template; VHome Uncertain items query hidden physical facts as of $t_q$.}
\label{tab:sim-qa-rules}
\footnotesize
\setlength{\tabcolsep}{3.5pt}
\renewcommand{\arraystretch}{1.08}
\begin{tabularx}{\textwidth}{@{}l>{\raggedright\arraybackslash}X>{\raggedright\arraybackslash}Xcc@{}}
\toprule
\rowcolor{stormhead!55}
Type & Known (THOR; VHome) & Uncertain & THOR K/U & VHome K/U \\
\midrule
Current state & Scene/object visibility \& state & Hidden physical state (VHome) & 46/0 & 52/1 \\
Factual retrieval & Scene/object visibility \& state & State during missing obs.\ (VHome) & 39/0 & 34/2 \\
History aggregation & Last return; return/exit counts & \# unobserved actions (VHome) & 35/0 & 26/12 \\
State change & Appearance / reviewed state seq. & -- & 43/0 & 34/0 \\
Object tracking & Returning category / reviewed state & Object state across a gap (VHome) & 24/0 & 8/23 \\
Temporal reasoning & Return / reviewed state order & Reapp.\ order (THOR); state order across gap (VHome) & 73/49 & 6/12 \\
\midrule
\rowcolor{stormhead!33}
Total & & & 260/49 & 160/50 \\
\bottomrule
\end{tabularx}
\end{table*}

\paragraph{Query-time paired items.}
\label{app:sim-thor-paired}
All THOR Uncertain items use one shared template with Known items: ``Which happens first: the X reappears at its original position or the Y reappears at its original position?'', with options X first, Y first, and both at the same time. Only objects that disappear and reappear once are used; both have disappeared by $t_q$. The order is Known when one object is seen back at its place and a later frame up to $t_q$ still shows the other place empty; otherwise the item is Uncertain (missing observation). Candidates are preselected from the camera plan and verified on frames (Appendix~\ref{sec:audit}); each pair is used at most once per label, and verified Known items pass the text-prior filter. Each 2\,s query-time bin keeps equal Known and Uncertain counts (49 each; AUROC 0.50 for query time alone).

\paragraph{Choices and natural-language realization.}
Each THOR item has three substantive options plus the shared abstention option in position D. A VLM refines fixed templates before export; the generator then samples lead-in and ending templates within each family (Appendix~\ref{app:sim-template-refinement}). Question text omits absolute timestamps. Evaluation permutes the three substantive options with a fixed seed; prompts hold the question and four options, selecting frames by~$t_q$.

\paragraph{Diagnostic labels.}
For THOR, let $s_j$ be the stored start of event $j$, obtained by rounding its planned before-observation time to two decimal places. The release computes
\begin{equation}
c_i^{\mathrm{sim}}=\max\!\left(1,\sum_j\mathbf{1}[s_j\leq t_{q,i}]\right).
\label{eq:sim-intensity}
\end{equation}
This counts event intervals from their before-observation boundary, describing accumulated change intensity at query time. The common bins are $1$--$3$, $4$--$6$, and $7$--$10$. Status labels select the corresponding question builder. Each Uncertain item is annotated with ground-truth uncertainty sources; all 49 THOR Uncertain items are query-time paired items labeled missing observation.

\subsubsection{Release validation and examples}
\label{app:sim-validation}

\paragraph{Structural and language checks.}
Release checks require unique IDs, three distinct substantive options, query and evidence times on the 1~FPS grid, positive-length intervals within $[0,t_q]$, shared abstention on every item, and Uncertain answers selecting it. Mean evidence-window length is 16.20\,s (41.4\% $>10$\,s). Questions omit numeric timestamps, and reappearance-order items share one stem.

\paragraph{Paired example.}
In THOR episode 12 (Figure~\ref{fig:case-sim}), the plate site is empty by 11\,s and the ladle site by 23\,s. At $t_q=41$\,s the ladle is back at 40\,s while the plate location has not been seen after that; at $t_q=44$\,s the 43\,s frame shows that location still empty, so the ladle returned first.

\paragraph{Reproduction artifacts.}
Released artifacts include room configs, scene/seed lists, plans, traces, validation records, and \texttt{questions.jsonl} with separate model-input and label files.

\subsubsection{VHome: native actions and grounded QA}
\label{app:sim-virtualhome}

\paragraph{Native room rollouts.}
VHome uses VirtualHome~\citep{puig2018virtualhome} to capture 29 room instances across seven native scenes: seven kitchens, seven living rooms, eight bedrooms, and seven bathrooms. Each accepted video contains ten scheduled events involving three to five distinct targets. Native probes verify reachable viewpoints, reversible operations, and uniquely describable objects before freezing the seeded program. One observer moves through the room while one or two separate characters execute native Open/Close, SwitchOn/SwitchOff, and Grab/PutBack operations. Programs include at least four distinct operations and five intended off-screen events; an off-screen manipulation must hide both its target and operator throughout the action. Accepted captures retain RGB frames, instance masks, event records, and graph-state transitions at $640\times480$ and 20~FPS. Encoded durations average 66.91\,s. Simulator success establishes that an action occurred; question answerability is determined from the sampled observations.

\paragraph{Evidence reconstruction and semantic scope.}
QA construction samples masks at 1~FPS. A target is visible at $\geq128$ pixels, absent at zero, and ambiguous between; transitions require two consecutive samples. Builders derive visibility, returns, order, and return/exit counts from these records. Physical-state candidates additionally use approved RGB endpoint pairs; hidden state, action count/order, and identity remain Uncertain when the prefix is insufficient. Release counts are 160 Known and 50 Uncertain (23.8\% Uncertain).

\paragraph{Trace-grounded Known items.}
\label{app:sim-vhome-trace}
Known items share templates with Uncertain ones and are grounded in the event log and masks. Count items require full visibility ($\geq128$ pixels) throughout $[t_1,t_2]$, endpoints at least 1\,s from any action, and every counted action fully observed. Single- and two-time state items require a visible, idle object at the queried time; two-time option sets are balanced across labels. Table~\ref{tab:sim-qa-rules} lists per-type counts.

\paragraph{Curation, wording, and validation.}
A deterministic curator selects ten items per episode across the six types with type/status and semantic-anchor controls. VLM wording edits preserve fact, temporal scope, evidence, and label (Appendix~\ref{app:sim-template-refinement}). Every item has three substantive options plus shared abstention; evaluation permutes the substantive options. All 50 Uncertain items record missing observation (12 also multiple candidates). Mean evidence length is 9.27\,s (31.0\% $>10$\,s). Intensity follows Eq.~\ref{eq:sim-intensity} with recorded event starts. Release checks cover 29 videos, unique IDs, schema fields, and causal evidence boundaries; model inputs and labels are stored separately.

%% file: tables/tab_qa_types.tex
\begin{table}[htbp]
\centering
\caption{\textbf{Semantic question types in the real-world track.} Each type targets a temporal reasoning operation, annotated independently from accumulated change intensity and Known/Uncertain status. Table~\ref{tab:sim-qa-rules} lists the corresponding simulation builders.}
\label{tab:qa-types}
\footnotesize
\renewcommand{\arraystretch}{1.08}
\setlength{\tabcolsep}{4pt}
\begin{tabularx}{\linewidth}{@{}
  >{\columncolor{stormhead!35}\raggedright\arraybackslash\bfseries}p{0.17\linewidth}
  >{\raggedright\arraybackslash}p{0.40\linewidth}
  >{\raggedright\arraybackslash}X@{}}
\toprule
\rowcolor{stormhead!55}
Type & Operational definition & Example question \\
\midrule

Factual retrieval &
Retrieve an object, attribute, or location from an earlier principal interval after the target leaves the view. &
\textit{``What color was the towel previously seen on the radiator?''} \\
\addlinespace[3pt]

Current state &
Find an entity's latest state or location from history up to $t_q$. &
\textit{``Where is the spice bottle at $t_q$ after the observed relocations?''} \\
\addlinespace[3pt]

State change &
Compare evidence across a transition to identify the state or placement change. &
\textit{``How did the milk frother's state change between 4\,s and 7\,s?''} \\
\addlinespace[3pt]

Object tracking &
Match timed views to one instance, or abstain. &
\textit{``Which earlier bowl, if any, is the bowl observed after the transition?''} \\
\addlinespace[3pt]

Temporal reasoning &
Order events or pick first/last/$k$-th actions. &
\textit{``Which occurred first: rinsing the knife, opening the drawer, or placing the bowl?''} \\
\addlinespace[3pt]

\makecell[tl]{History\\aggregation} &
Count or summarize visible action segments up to~$t_q$. &
\textit{``How many distinct visible drawer-opening segments occurred by $t_q$?''} \\

\bottomrule
\end{tabularx}
\end{table}

%% file: tables/tab_main_stats.tex
\begin{table*}[t]
\centering
\caption{\textbf{\bench{} per-subset statistics.} Source, episodes, questions, total hours, mean duration, QA density, Known/Uncertain counts, Low/Medium/High change-intensity bins ($1$--$3$/$4$--$6$/$7$--$10$), and the Earlier share whose evidence starts before the latest 16 frames before $t_q$. Complements the composition figure in the main paper. The Total row aggregates counts across subsets; duration, QA density, and Earlier share are computed over the pooled release.}
\label{tab:main-stats}
\scriptsize
\setlength{\tabcolsep}{3.2pt}
\renewcommand{\arraystretch}{1.12}
\begin{tabularx}{\textwidth}{@{}ll*{8}{>{\centering\arraybackslash}X}@{}}
\toprule
\rowcolor{stormhead!55}
\textbf{Subset} & \textbf{Source} & \textbf{Ep.} & \textbf{QA} & \textbf{Hours} & \textbf{Dur.} & \textbf{QA/min} & \textbf{Known/Unc.} & \textbf{L/M/H} & \textbf{Earlier} \\
\midrule
Cook & HD-EPIC & 307 & 990 & 3.649 & 42.8\,s & 4.5 & 711/279 & 280/424/286 & 48.8\% \\
Bike & Ego-Exo4D & 61 & 389 & 0.661 & 39.0\,s & 9.8 & 293/96 & 195/174/20 & 43.7\% \\
Health & Ego-Exo4D & 82 & 297 & 0.936 & 41.1\,s & 5.3 & 209/88 & 113/129/55 & 37.0\% \\
Music & Ego-Exo4D & 16 & 76 & 0.186 & 41.7\,s & 6.8 & 53/23 & 45/31/0 & 57.9\% \\
Sports & Ego-Exo4D & 99 & 306 & 1.125 & 40.9\,s & 4.5 & 216/90 & 154/111/41 & 29.7\% \\
THOR & AI2-THOR & 28 & 309 & 0.467 & 60.1\,s & 11.0 & 260/49 & 22/53/234 & 38.2\% \\
VHome & VirtualHome & 29 & 210 & 0.539 & 66.9\,s & 6.5 & 160/50 & 43/52/115 & 32.9\% \\
\midrule
\rowcolor{stormhead!33}
\textbf{Total} & & \textbf{622} & \textbf{2,577} & \textbf{7.563} & \textbf{43.8\,s} & \textbf{5.7} & \textbf{1,902/675} & \textbf{852/974/751} & \textbf{42.1\%} \\
\bottomrule
\end{tabularx}
\end{table*}

%% file: tables/tab_dataset_diagnostics.tex
\begin{table*}[t]
\centering
\caption{\textbf{Benchmark structural composition.} Left: question-type counts (Cur.\ current; Fact.\ factual; Hist.\ history; Chg.\ change; Track; Temp.). Right: Uncertain-source counts (Miss / Part / AmbE / LowQ / Multi / AmbA). Bottom: mean change intensity, query time, and evidence-span diagnostics (Earlier share: Table~\ref{tab:main-stats}).}
\label{tab:dataset-diagnostics}
\scriptsize
\setlength{\tabcolsep}{3.2pt}
\renewcommand{\arraystretch}{1.08}
\begin{minipage}[t]{0.48\textwidth}
\centering
\textbf{Question-type counts}\\[0.35em]
\begin{tabularx}{\linewidth}{@{}l*{7}{>{\centering\arraybackslash}X}@{}}
\toprule
\rowcolor{stormhead!55}
Domain & Cur. & Fact. & Hist. & Chg. & Track & Temp. & All \\
\midrule
Cook & 174 & 176 & 147 & 152 & 162 & 179 & 990 \\
Bike & 79 & 80 & 57 & 69 & 60 & 44 & 389 \\
Health & 61 & 82 & 33 & 51 & 48 & 22 & 297 \\
Music & 14 & 19 & 10 & 14 & 15 & 4 & 76 \\
Sports & 78 & 83 & 26 & 44 & 56 & 19 & 306 \\
THOR & 46 & 39 & 35 & 43 & 24 & 122 & 309 \\
VHome & 53 & 36 & 38 & 34 & 31 & 18 & 210 \\
\midrule
\rowcolor{stormhead!33}
Total & 505 & 515 & 346 & 407 & 396 & 408 & 2,577 \\
\bottomrule
\end{tabularx}
\end{minipage}
\hfill
\begin{minipage}[t]{0.48\textwidth}
\centering
\textbf{Uncertainty-source frequencies}\\[0.35em]
\begin{tabularx}{\linewidth}{@{}l*{6}{>{\centering\arraybackslash}X}@{}}
\toprule
\rowcolor{stormhead!55}
Domain & Miss & Part & AmbE & LowQ & Multi & AmbA \\
\midrule
Cook & 224 & 105 & 63 & 39 & 12 & 7 \\
Bike & 69 & 52 & 15 & 9 & 20 & 6 \\
Health & 48 & 59 & 27 & 26 & 8 & 11 \\
Music & 11 & 14 & 7 & 8 & 3 & 4 \\
Sports & 57 & 49 & 22 & 24 & 20 & 17 \\
THOR & 49 & 0 & 0 & 0 & 0 & 0 \\
VHome & 50 & 0 & 0 & 0 & 12 & 0 \\
\midrule
\rowcolor{stormhead!33}
Total & 508 & 279 & 134 & 106 & 75 & 45 \\
\bottomrule
\end{tabularx}
\end{minipage}

\vspace{0.85em}
\centering
\textbf{Change intensity, query time, and evidence spans}\\[0.35em]
\setlength{\tabcolsep}{2.8pt}
\begin{tabularx}{\textwidth}{@{}l*{8}{>{\centering\arraybackslash}X}@{}}
\toprule
\rowcolor{stormhead!55}
Domain & Int. & K / U & $t_q$ (s) & $t_q$ pos. & Spans & Instant & Mean dur. & $>$10\,s \\
\midrule
Cook & 5.11 & 4.68 / 6.21 & 29.66 & 69.7\% & 2.70 & 73.0\% & 5.26 & 7.5\% \\
Bike & 3.61 & 3.52 / 3.91 & 27.18 & 70.1\% & 2.02 & 55.9\% & 4.91 & 9.8\% \\
Health & 4.42 & 3.89 / 5.68 & 26.47 & 65.1\% & 2.95 & 86.0\% & 5.50 & 5.4\% \\
Music & 3.16 & 2.94 / 3.65 & 27.78 & 68.3\% & 4.24 & 90.4\% & 12.23 & 13.2\% \\
Sports & 3.81 & 3.26 / 5.12 & 24.98 & 62.1\% & 2.58 & 89.1\% & 3.78 & 1.6\% \\
THOR & 7.34 & 7.40 / 7.02 & 42.85 & 71.4\% & 1.00 & 0.0\% & 16.20 & 41.4\% \\
VHome & 6.43 & 6.76 / 5.36 & 40.98 & 61.3\% & 1.00 & 0.0\% & 9.27 & 31.0\% \\
\bottomrule
\end{tabularx}
\end{table*}

%% file: tables/tab_overview_density.tex
\begin{table}[htbp]
\centering
\caption{\textbf{Duration and QA density behind Figure~\ref{fig:teaser}b.} Durations use local media ($n$ videos; median in seconds). QA density is QA count divided by total video duration.}
\label{tab:overview-density}
\scriptsize
\setlength{\tabcolsep}{3pt}
\renewcommand{\arraystretch}{1.05}
\begin{tabularx}{\columnwidth}{@{}>{\raggedright\arraybackslash}Xrrr@{}}
\toprule
\rowcolor{stormhead!55}
Dataset & Duration $n$ & Median (s) & QA/min \\
\midrule
EgoSchema & 1,157 & 180.00 & 0.33 \\
LongVideoBench & 3,991 & 60.80 & 0.23 \\
LVBench & 103 & 3664.41 & 0.22 \\
MLVU & 2,106 & 480.02 & 0.12 \\
Video-MME & 900 & 487.94 & 0.18 \\
\rowcolor{stormhead!60}
\bench{} & 622 & 44.80 & 5.68 \\
\bottomrule
\end{tabularx}
\end{table}

%% file: tables/tab_frame_budget.tex
\begin{table}[H]
\caption{\tabtitle{Per-model frame budget under the 1~FPS online protocol (RQ1).} Checkpoints share the 1~FPS clock and retain recent frames (\emph{Max}) up to native caps, padding the final frame to temporal unit (\emph{Unit}). Episodes last under 70~s, so only the 32-frame cap of LLaVA-NeXT-Video can truncate the history; the 16-frame control in Table~\ref{tab:input-controls} tests a shorter window. MiniCPM-V~4.5 packs as many as three frames into each temporal group.}
\label{tab:frame-budget}
\centering
\scriptsize
\setlength{\tabcolsep}{3pt}
\renewcommand{\arraystretch}{0.98}
\begin{tabularx}{\columnwidth}{@{}>{\raggedright\arraybackslash}Xlc r@{}}
\toprule
\rowcolor{stormhead!55}
\tabhead{Model} & \tabhead{Family} & \tabhead{Max} & \tabhead{Unit} \\
\midrule
Qwen2.5-VL-3B & Qwen & 768 & 2 \\
Qwen2.5-VL-7B & Qwen & 768 & 2 \\
Qwen3-VL-4B & Qwen & 768 & 2 \\
Qwen3-VL-8B & Qwen & 768 & 2 \\
Qwen3-VL-32B & Qwen & 768 & 2 \\
Qwen3.5-4B & Qwen & 768 & 2 \\
Qwen3.5-9B & Qwen & 768 & 2 \\
InternVL3.5-8B & InternVL & 240 & 1 \\
Molmo2-8B & Molmo2 & 384 & 1 \\
MiniCPM-V~4.5 & MiniCPM & 180 & 1 (pack 3) \\
InternVideo2.5-8B & InternVideo & 512 & 4 \\
Eagle2.5-8B & Eagle & 256 & 2 \\
VideoLLaMA3-7B & VideoLLaMA3 & 128 & 2 \\
LLaVA-NeXT-Video-7B-DPO & LLaVA-NeXT-Video & \cellcolor{stormhead}32 & 1 \\
GLM-4.1V-9B-Thinking & GLM-4.1V & 240 & 2 \\
\bottomrule
\end{tabularx}
\end{table}

%% file: tables/tab_streaming_models.tex
\begin{table}[H]
\centering
\caption{\textbf{Streaming models in native mode.} All models receive the same 1\,FPS frames once as they become visible, keep state across queries within an episode, and reset between episodes; questions and answers are never written back into state.}
\label{tab:streaming-models}
\scriptsize
\setlength{\tabcolsep}{2.5pt}
\renewcommand{\arraystretch}{0.98}
\begin{tabularx}{\columnwidth}{@{}>{\raggedright\arraybackslash}p{1.55cm}>{\raggedright\arraybackslash}p{1.35cm}>{\raggedright\arraybackslash}X@{}}
\toprule
\rowcolor{stormhead!55}
Model & Backbone & State / setting in \bench{} \\
\midrule
Flash-VStream~\citep{zhang2025flashvstream} & Qwen2-VL-7B & Flash Memory (1 frame/clip) \\
LiveCC~\citep{chen2025livecc} & Qwen2-VL-7B & KV cache; 1 chat turn / 2-frame clip; 1\,FPS (was 2\,FPS); silent per-clip replies \\
TimeChat-Online~\citep{yao2025timechatonline} & Qwen2.5-VL-7B & Diff.\ token drop (thr.\ 0.5), 2-frame step; cache kept tokens (no prefix re-encode) \\
ReKV~\citep{di2025rekv} & LLaVA-OV-7B & Video KV + question-guided retrieval; 1\,FPS (was 0.5\,FPS) \\
MovieChat~\citep{song2024moviechat} & Vicuna-7B & Short/long-term memory, 8-frame fragment; breakpoint mode; incomplete fragment on a memory copy \\
StreamChat\linebreak[0]~\citep{xiong2025streamchat} & LongVA-7B & Hierarchical memory; captioned leaf / 40 frames; 1\,FPS; incomplete chunk as temporary leaf \\
Dispider~\citep{qian2025dispider} & Qwen2-7B & Perception memory; 1 entry / 16\,s clip; no response trigger; open clip perceived at $t_q$ \\
\bottomrule
\end{tabularx}
\end{table}

%% file: tables/tab_rq2_offline_br.tex
\begin{table}[H]
\centering
\caption{\textbf{Full-video vs.\ causal-prefix (RQ2).} Seven-subset $\KA$/$\AR$/$\HM$ (\%) for Full vs.\ Online ($V_{\le t_q}$). $\Delta$ denotes the paired per-model difference (Online$-$Full); the Median row is the median of those paired deltas.}
\label{tab:rq2-offline-br}
\scriptsize
\setlength{\tabcolsep}{2.2pt}
\renewcommand{\arraystretch}{1.05}
\begin{tabularx}{\columnwidth}{@{}>{\raggedright\arraybackslash}X*{3}{rr>{\columncolor{stormhead!35}}r}@{}}
\toprule
\rowcolor{stormhead!55}
& \multicolumn{3}{c}{$\KA$} & \multicolumn{3}{c}{$\AR$} & \multicolumn{3}{c}{$\HM$} \\
\rowcolor{stormhead!55}
Model & Full & On. & $\Delta$ & Full & On. & $\Delta$ & Full & On. & $\Delta$ \\
\midrule
Q2.5-VL-3B & 33.3 & 32.5 & -0.7 & 47.0 & 47.5 & +0.5 & 36.1 & 36.1 & +0.1 \\
Q2.5-VL-7B & 30.0 & 31.2 & +1.2 & 77.2 & 80.0 & +2.8 & 41.1 & 42.8 & +1.7 \\
Q3-VL-8B & 63.9 & 65.5 & +1.6 & 16.9 & 19.1 & +2.1 & 24.4 & 26.8 & +2.4 \\
IVL3.5-8B & 63.6 & 66.0 & +2.4 & 8.9 & 8.7 & -0.2 & 14.9 & 14.6 & -0.3 \\
MCPM-V 4.5 & 63.6 & 63.5 & -0.1 & 18.5 & 22.7 & +4.2 & 27.5 & 32.0 & +4.6 \\
\midrule
\rowcolor{stormhead!33}
Median & 63.6 & 63.5 & +1.2 & 18.5 & 22.7 & +2.1 & 27.5 & 32.0 & +1.7 \\
\bottomrule
\end{tabularx}
\end{table}

%% file: tables/tab_diagnostic_cells.tex
\begin{table}[H]
\centering
\caption{\textbf{Six-cell decomposition on Cook (RQ3).} $\KA$/$\AR$ (\%) by change-intensity bin; Lowest cell and FAR on Known.}
\label{tab:diagnostic-cells}
\scriptsize
\setlength{\tabcolsep}{1.6pt}
\renewcommand{\arraystretch}{1.02}
\begin{tabularx}{\columnwidth}{@{}>{\raggedright\arraybackslash}X*{8}{r}@{}}
\toprule
\rowcolor{stormhead!55}
& \multicolumn{2}{c}{Low} & \multicolumn{2}{c}{Med.} & \multicolumn{2}{c}{High} & \multicolumn{2}{c@{}}{\cellcolor{stormhead}Sum.} \\
\rowcolor{stormhead!55}
Model & $\KA$ & $\AR$ & $\KA$ & $\AR$ & $\KA$ & $\AR$ & \cellcolor{stormhead}Low. & \cellcolor{stormhead}FAR \\
\midrule
Q2.5-VL-3B & 35.5 & 41.4 & 38.7 & 54.3 & 35.0 & 57.7 & 35.0 & 29.0 \\
Q2.5-VL-7B & 41.4 & 75.9 & 40.4 & 71.7 & 35.0 & 74.0 & 35.0 & 38.4 \\
Q3-VL-4B & 74.9 & 31.0 & 68.0 & 28.3 & 67.5 & 26.8 & 26.8 & 4.9 \\
Q3-VL-8B & 79.3 & 20.7 & 67.0 & 12.6 & 71.2 & 13.8 & 12.6 & 0.8 \\
Q3-VL-32B & 75.3 & 34.5 & 74.7 & 35.4 & 75.5 & 31.7 & 31.7 & 2.8 \\
Q3.5-4B & 74.1 & 44.8 & 68.7 & 44.9 & 69.3 & 42.3 & 42.3 & 9.6 \\
Q3.5-9B & 86.1 & 20.7 & 77.4 & 33.9 & 84.7 & 24.4 & 20.7 & 3.1 \\
IVL3.5-8B & 75.3 & 10.3 & 72.7 & 11.0 & 77.9 & 7.3 & 7.3 & 1.0 \\
Molmo2-8B & 74.9 & 37.9 & 65.7 & 52.0 & 66.3 & 54.5 & 37.9 & 15.8 \\
MCPM-V 4.5 & 77.7 & 31.0 & 73.7 & 27.6 & 74.2 & 16.3 & 16.3 & 1.1 \\
IVideo2.5-8B & 64.1 & 13.8 & 62.3 & 16.5 & 65.0 & 18.7 & 13.8 & 4.1 \\
Eagle2.5-8B & 81.3 & 17.2 & 72.7 & 30.7 & 78.5 & 26.8 & 17.2 & 2.3 \\
VLLaMA3-7B & 76.5 & 44.8 & 64.3 & 47.2 & 65.6 & 49.6 & 44.8 & 12.7 \\
LLaVA-NeXT-V & 40.2 & 3.4 & 41.8 & 3.9 & 39.3 & 3.3 & 3.3 & 1.4 \\
GLM-4.1V & 51.4 & 69.0 & 42.4 & 63.0 & 42.3 & 58.5 & 42.3 & 29.3 \\
\midrule
\rowcolor{stormhead!68}
\multicolumn{9}{@{}l}{\textit{Streaming (native state)}} \\
Flash-VStream & 63.3 & 24.1 & 47.1 & 28.3 & 44.2 & 35.0 & 24.1 & 12.5 \\
LiveCC & 41.0 & 72.4 & 37.0 & 63.0 & 35.0 & 63.4 & 35.0 & 46.7 \\
TimeChat-Online & 64.9 & 34.5 & 60.9 & 37.0 & 60.1 & 30.1 & 30.1 & 9.6 \\
ReKV & 68.5 & 13.8 & 57.6 & 19.7 & 58.3 & 26.8 & 13.8 & 4.9 \\
MovieChat & 35.1 & 10.3 & 33.0 & 17.3 & 38.7 & 15.4 & 10.3 & 13.1 \\
StreamChat & 33.1 & 65.5 & 32.0 & 51.2 & 25.8 & 50.4 & 25.8 & 42.3 \\
Dispider & 53.4 & 48.3 & 50.2 & 52.0 & 48.5 & 52.8 & 48.3 & 25.0 \\
\bottomrule
\end{tabularx}
\end{table}

%% file: tables/tab_sim_results.tex
\begin{table}[H]
\centering
\caption{\textbf{Simulation results (THOR / VHome).} $\KA$, $\AR$, $\HM$ with 95\% episode-clustered CI, and FAR (\%).}
\label{tab:sim-results}
\scriptsize
\setlength{\tabcolsep}{1.4pt}
\renewcommand{\arraystretch}{1.02}
\begin{tabularx}{\columnwidth}{@{}>{\raggedright\arraybackslash}X*{2}{cc>{\columncolor{stormhead!35}}cc}@{}}
\toprule
\rowcolor{stormhead!55}
& \multicolumn{4}{c}{THOR} & \multicolumn{4}{c}{VHome} \\
\rowcolor{stormhead!55}
Model & $\KA$ & $\AR$ & \cellcolor{stormhead}$\HM$ & FAR & $\KA$ & $\AR$ & \cellcolor{stormhead}$\HM$ & FAR \\
\midrule
Q2.5-VL-3B & 46.5 & 32.7 & 38.4{\tiny[28,47]} & 23.8 & 26.2 & 12.0 & 16.5{\tiny[7,23]} & 14.4 \\
Q2.5-VL-7B & 31.2 & 89.8 & 46.3{\tiny[41,51]} & 60.0 & 8.8 & 90.0 & 15.9{\tiny[8,24]} & 81.2 \\
Q3-VL-4B & 56.9 & 10.2 & 17.3{\tiny[5,29]} & 5.0 & 53.1 & 8.0 & 13.9{\tiny[4,25]} & 0.6 \\
Q3-VL-8B & 56.5 & 0.0 & 0.0{\tiny[0,0]} & 0.4 & 49.4 & 8.0 & 13.8{\tiny[4,24]} & 1.2 \\
Q3-VL-32B & 57.7 & 6.1 & 11.1{\tiny[0,22]} & 4.2 & 47.5 & 18.0 & 26.1{\tiny[13,35]} & 3.1 \\
Q3.5-4B & 53.8 & 34.7 & 42.2{\tiny[32,51]} & 21.5 & 55.0 & 28.0 & 37.1{\tiny[22,48]} & 6.2 \\
Q3.5-9B & 58.1 & 0.0 & 0.0{\tiny[0,0]} & 11.5 & 58.1 & 14.0 & 22.6{\tiny[8,34]} & 2.5 \\
IVL3.5-8B & 58.1 & 0.0 & 0.0{\tiny[0,0]} & 0.8 & 53.8 & 4.0 & 7.4{\tiny[0,17]} & 0.6 \\
Molmo2-8B & 55.0 & 12.2 & 20.0{\tiny[4,33]} & 6.2 & 33.1 & 64.0 & 43.7{\tiny[37,50]} & 48.8 \\
MCPM-V 4.5 & 48.8 & 28.6 & 36.1{\tiny[24,45]} & 18.8 & 45.6 & 4.0 & 7.4{\tiny[0,16]} & 6.9 \\
IVideo2.5-8B & 56.2 & 4.1 & 7.6{\tiny[0,17]} & 8.5 & 48.1 & 20.0 & 28.3{\tiny[16,39]} & 11.9 \\
Eagle2.5-8B & 41.2 & 18.4 & 25.4{\tiny[12,34]} & 36.2 & 36.2 & 48.0 & 41.3{\tiny[34,47]} & 41.2 \\
VLLaMA3-7B & 38.8 & 14.3 & 20.9{\tiny[8,31]} & 38.5 & 49.4 & 34.0 & 40.3{\tiny[28,49]} & 18.1 \\
LLaVA-NeXT-V & 51.9 & 0.0 & 0.0{\tiny[0,0]} & 0.8 & 33.1 & 0.0 & 0.0{\tiny[0,0]} & 0.0 \\
GLM-4.1V & 32.3 & 100 & 48.8{\tiny[45,53]} & 51.2 & 39.4 & 76.0 & 51.9{\tiny[44,59]} & 23.8 \\
\midrule
\rowcolor{stormhead!68}
\multicolumn{9}{@{}l}{\textit{Streaming (native state)}} \\
Flash-VStream & 45.0 & 18.4 & 26.1{\tiny[15,34]} & 11.5 & 15.0 & 22.0 & 17.8{\tiny[11,23]} & 60.0 \\
LiveCC & 42.7 & 4.1 & 7.5{\tiny[0,17]} & 35.0 & 21.2 & 26.0 & 23.4{\tiny[15,29]} & 26.2 \\
TimeChat-Online & 54.6 & 2.0 & 3.9{\tiny[0,11]} & 16.9 & 33.1 & 18.0 & 23.3{\tiny[11,30]} & 25.6 \\
ReKV & 56.5 & 4.1 & 7.6{\tiny[0,17]} & 11.5 & 40.6 & 8.0 & 13.4{\tiny[4,23]} & 16.9 \\
MovieChat & 37.7 & 8.2 & 13.4{\tiny[4,23]} & 14.2 & 36.2 & 16.0 & 22.2{\tiny[13,29]} & 15.6 \\
StreamChat & 30.8 & 6.1 & 10.2{\tiny[0,18]} & 55.0 & 16.2 & 62.0 & 25.8{\tiny[18,33]} & 51.9 \\
Dispider & 45.4 & 0.0 & 0.0{\tiny[0,0]} & 26.9 & 38.1 & 42.0 & 40.0{\tiny[33,46]} & 25.0 \\
\bottomrule
\end{tabularx}
\end{table}

%% file: tables/tab_global_diagnostics.tex
\begin{table}[H]
\centering
\caption{\textbf{Seven-subset diagnostics.} $\KA$/$\AR$ (\%) averages; $B_1$/$B_3$\,=\,Low/High change; $\Delta$Earlier\,=\,earlier minus Last-16.}
\label{tab:global-diagnostics}
\scriptsize
\setlength{\tabcolsep}{1.3pt}
\renewcommand{\arraystretch}{1.02}
\begin{tabularx}{\columnwidth}{@{}>{\raggedright\arraybackslash}X*{7}{r}>{\columncolor{stormhead!35}}r>{\columncolor{stormhead!35}}r@{}}
\toprule
\rowcolor{stormhead!55}
Model & $\KA$ & $\AR$ & $\KA{-}\AR$ & $B_1$ & $B_3$ & Cur. & Hist. & $\Delta$K & $\Delta$A \\
\midrule
Q2.5-VL-3B & 32.5 & 47.5 & -14.9 & 32.4 & 29.5 & 35.7 & 27.9 & -3.9 & +4.9 \\
Q2.5-VL-7B & 31.2 & 80.0 & -48.9 & 30.6 & 31.4 & 35.0 & 27.9 & -1.9 & -2.4 \\
Q3-VL-4B & 67.1 & 28.5 & +38.6 & 62.7 & 59.3 & 68.8 & 59.4 & +0.4 & -7.6 \\
Q3-VL-8B & 65.5 & 19.1 & +46.4 & 63.9 & 65.7 & 63.7 & 52.3 & +0.3 & -5.0 \\
Q3-VL-32B & 67.7 & 32.1 & +35.7 & 63.6 & 65.6 & 73.0 & 56.6 & -1.1 & -5.2 \\
Q3.5-4B & 67.7 & 40.6 & +27.1 & 63.6 & 65.9 & 66.2 & 49.1 & +3.6 & -5.9 \\
Q3.5-9B & 75.8 & 25.0 & +50.8 & 72.6 & 71.1 & 79.8 & 57.2 & +3.9 & -5.5 \\
IVL3.5-8B & 66.0 & 8.7 & +57.3 & 62.3 & 63.0 & 65.7 & 53.0 & +0.1 & -1.3 \\
Molmo2-8B & 51.0 & 60.3 & -9.3 & 51.7 & 50.9 & 48.8 & 34.0 & +0.6 & -15.1 \\
MCPM-V 4.5 & 63.5 & 22.7 & +40.8 & 60.7 & 62.2 & 61.8 & 56.8 & +1.3 & +0.7 \\
IVideo2.5-8B & 60.6 & 23.8 & +36.8 & 59.3 & 61.0 & 59.5 & 57.8 & +2.1 & -11.1 \\
Eagle2.5-8B & 62.4 & 36.7 & +25.6 & 61.7 & 60.9 & 63.0 & 54.7 & +1.9 & +1.9 \\
VLLaMA3-7B & 57.8 & 47.4 & +10.4 & 58.8 & 50.7 & 63.8 & 43.3 & -6.6 & -5.4 \\
LLaVA-NeXT-V & 38.4 & 2.9 & +35.5 & 34.5 & 37.3 & 42.2 & 33.4 & +0.2 & +0.4 \\
GLM-4.1V & 42.2 & 72.1 & -29.9 & 42.1 & 41.9 & 44.5 & 36.3 & -4.1 & +1.4 \\
\midrule
\rowcolor{stormhead!68}
\multicolumn{10}{@{}l}{\textit{Streaming (native state)}} \\
Flash-VStream & 45.6 & 23.7 & +21.9 & 49.8 & 38.7 & 47.8 & 36.9 & -8.7 & +0.2 \\
LiveCC & 36.8 & 41.7 & -5.0 & 34.9 & 37.8 & 38.2 & 15.5 & +0.5 & +1.0 \\
TimeChat-Online & 55.0 & 28.6 & +26.4 & 48.9 & 56.3 & 58.6 & 49.2 & +0.0 & -8.3 \\
ReKV & 56.1 & 17.6 & +38.5 & 53.8 & 54.5 & 57.4 & 43.2 & -1.0 & +1.4 \\
MovieChat & 36.6 & 11.9 & +24.6 & 33.4 & 36.3 & 41.4 & 36.0 & -0.6 & -1.2 \\
StreamChat & 30.9 & 39.0 & -8.0 & 27.1 & 33.4 & 31.8 & 13.2 & +0.4 & -6.6 \\
Dispider & 46.6 & 35.3 & +11.2 & 47.1 & 46.6 & 50.1 & 34.5 & +0.6 & +0.9 \\
\midrule
\rowcolor{stormhead!33}
Median & 55.5 & 30.3 & +26.0 & 52.8 & 52.7 & 58.0 & 43.3 & +0.2 & -1.9 \\
\bottomrule
\end{tabularx}
\end{table}

%% file: tables/tab_rq3_sources.tex
\begin{table}[H]
\centering
\caption{\textbf{Abstention recall by uncertainty source (RQ3).} Item counts $n$ (multi-label) and median $\AR$ (\%) over 22 models.}
\label{tab:rq3-sources}
\scriptsize
\setlength{\tabcolsep}{4pt}
\renewcommand{\arraystretch}{1.05}
\begin{tabularx}{\columnwidth}{@{}>{\raggedright\arraybackslash}Xr>{\columncolor{stormhead!35}}r@{}}
\toprule
\rowcolor{stormhead!55}
Source & $n$ & Median $\AR$ \\
\midrule
\rowcolor{stormhead!33}
Missing observation & 508 & \cellcolor{stormhead}26.6 \\
Partial observation & 279 & 36.4 \\
Ambiguous evidence & 134 & 28.7 \\
\rowcolor{stormhead!33}
Low visual quality & 106 & \cellcolor{stormhead}62.3 \\
Multiple candidates & 75 & 18.7 \\
Ambiguous attribute & 45 & 44.4 \\
\bottomrule
\end{tabularx}
\end{table}

%% file: tables/tab_input_controls.tex
\begin{table}[H]
\centering
\caption{\textbf{Input controls (RQ4).} Seven-subset $\HM$ for five controls and median $\KA$/$\AR$/$\HM$ (\%). Forced-choice rows report $\KA$ only.}
\label{tab:input-controls}
\label{tab:cook-ablations}
\scriptsize
\setlength{\tabcolsep}{1.5pt}
\renewcommand{\arraystretch}{1.05}
\begin{tabularx}{\columnwidth}{@{}>{\raggedright\arraybackslash}X*{5}{r}>{\columncolor{stormhead!35}}r>{\columncolor{stormhead!35}}r>{\columncolor{stormhead!35}}r@{}}
\toprule
\rowcolor{stormhead!55}
& \multicolumn{5}{c}{$\HM$ per model} & \multicolumn{3}{c@{}}{\cellcolor{stormhead}Median} \\
\rowcolor{stormhead!55}
Input & 3B & 7B & Q3-8B & IVL & MCPM & \cellcolor{stormhead}$\KA$ & \cellcolor{stormhead}$\AR$ & \cellcolor{stormhead}$\HM$ \\
\midrule
Video $\le t_q$ & 36.1 & 42.8 & 26.8 & 14.6 & 32.0 & \cellcolor{stormhead}63.5 & 22.7 & 32.0 \\
Latest 16 frames & 36.9 & 40.1 & 29.8 & 17.1 & 41.5 & 60.9 & 34.6 & 36.9 \\
Shuffled frames & 33.3 & 36.8 & 25.7 & 14.3 & 31.7 & 57.4 & 22.8 & 31.7 \\
Black frames & 0.1 & 0.4 & 6.6 & 10.9 & 16.1 & \cellcolor{stormhead}3.5 & 83.0 & 6.6 \\
Text only & 1.9 & 0.3 & 25.3 & 16.6 & 21.8 & \cellcolor{stormhead}16.0 & 57.1 & 16.6 \\
Options only & 0.0 & 0.0 & 12.4 & 21.5 & 3.3 & \cellcolor{stormhead}1.7 & 96.7 & 3.3 \\
\midrule
\rowcolor{stormhead!68}
\multicolumn{9}{@{}l}{\textit{Forced choice ($\KA$; chance 33.3)}} \\
Text only & 23.9 & 21.8 & 29.1 & 27.9 & 32.2 & 27.9 & -- & -- \\
Video $\le t_q$ & 45.2 & 53.2 & 67.6 & 66.7 & 67.2 & 66.7 & -- & -- \\
\bottomrule
\end{tabularx}
\\[0.35em]
{\scriptsize 3B/7B\,=\,Qwen2.5-VL; Q3-8B\,=\,Qwen3-VL-8B; IVL\,=\,InternVL3.5-8B; MCPM\,=\,MiniCPM-V~4.5.}
\end{table}

%% file: tables/tab_stale_history.tex
\begin{table}[H]
\centering
\caption{\textbf{Valid vs.\ stale history (RQ4).} $\KA$/$\AR$ (\%) on full prefix and $\Delta$ under last-16 frames, by evidence window.}
\label{tab:stale-history}
\scriptsize
\setlength{\tabcolsep}{2.0pt}
\renewcommand{\arraystretch}{1.05}
\begin{tabularx}{\columnwidth}{@{}>{\raggedright\arraybackslash}X*{2}{rrr>{\columncolor{stormhead!35}}r}@{}}
\toprule
\rowcolor{stormhead!55}
& \multicolumn{4}{c}{$\KA$ (Known)} & \multicolumn{4}{c}{$\AR$ (Unc.)} \\
\rowcolor{stormhead!55}
& \multicolumn{2}{c}{Last16} & \multicolumn{2}{c}{\cellcolor{stormhead}Earlier} & \multicolumn{2}{c}{Last16} & \multicolumn{2}{c@{}}{\cellcolor{stormhead}Earlier} \\
\rowcolor{stormhead!55}
Model & Full & $\Delta$ & \cellcolor{stormhead}Full & \cellcolor{stormhead}$\Delta$ &
Full & $\Delta$ & \cellcolor{stormhead}Full & \cellcolor{stormhead}$\Delta$ \\
\midrule
Q2.5-VL-3B & 36.5 & -2.2 & 32.6 & -7.3 & 47.7 & +2.1 & 52.6 & +8.1 \\
Q2.5-VL-7B & 36.1 & -2.8 & 34.2 & -9.1 & 77.8 & +2.4 & 75.4 & +6.4 \\
Q3-VL-8B & 67.5 & -1.2 & 67.8 & -10.6 & 19.5 & +0.0 & 14.5 & +7.8 \\
IVL3.5-8B & 69.0 & +2.1 & 69.0 & -5.8 & 9.4 & +1.5 & 8.1 & +2.3 \\
MCPM-V 4.5 & 66.6 & +2.5 & 67.9 & -10.1 & 21.9 & +4.0 & 22.5 & +17.6 \\
\midrule
\rowcolor{stormhead!33}
Median & 66.6 & -1.2 & 67.8 & \cellcolor{stormhead}-9.1 & 21.9 & +2.1 & 22.5 & \cellcolor{stormhead}+7.8 \\
\bottomrule
\end{tabularx}
\end{table}

%% file: contents/appendix_prompts.tex
\lstdefinestyle{stormpromptlisting}{
  basicstyle=\ttfamily\scriptsize\color{stormdark},
  breaklines=true,
  breakatwhitespace=false,
  columns=fullflexible,
  keepspaces=true,
  showstringspaces=false,
  aboveskip=0pt,
  belowskip=0pt
}

\newtcblisting{stormprompt}[1]{
  enhanced,
  breakable,
  listing only,
  colback=stormhead!55,
  colframe=stormblue!55!stormdark,
  colbacktitle=stormhead,
  coltitle=stormdark,
  title={#1},
  fonttitle=\bfseries\small,
  boxrule=0.45pt,
  arc=1.2mm,
  outer arc=1.2mm,
  left=1.6mm,
  right=1.6mm,
  top=0.8mm,
  bottom=0.8mm,
  toptitle=0.7mm,
  bottomtitle=0.7mm,
  before skip=0.6em,
  after skip=0.8em,
  listing options={style=stormpromptlisting},
  lefttitle=1.2mm,
  righttitle=1.2mm
}

\section{Prompt Templates}
\label{sec:appendix-prompts}

\subsection{Real-world Cook/Kitchen prompts}
\label{app:kitchen_prompts}

The templates below are used to construct Cook/Kitchen episodes; braces denote fields filled at runtime.

\input{contents/appendix_kitchen_prompts}

\input{contents/appendix_sim_prompt_refinement}

%% file: contents/appendix_kitchen_prompts.tex
\begin{stormprompt}{Semantic Activity Segmentation}
<timestamped kitchen frames>
Processing interval: [{interval_start}, {interval_end})

You are the semantic activity segmentation assistant for RevisitQA, an embodied revisit benchmark in kitchen environments. You will receive uniformly sampled frames from a first-person kitchen video. Every frame is paired with a timestamp in seconds.

A kitchen contains several functional regions: washing occurs around the sink, food preparation occurs around a cutting board or worktop, cooking occurs around a stove and cookware, food is retrieved from or stored in a refrigerator, and utensils or condiments are retrieved from or stored in cabinets and drawers.

Divide the continuous video into semantically coherent atomic activity intervals. Base segmentation on the functional region, the dominant activity phase, and any clearly visible semantic outcome. Do not organize the timeline around a preselected target object.

Create a new boundary when, in priority order:
1. The functional region changes.
2. The dominant activity phase changes, for example: approach the sink -> clean -> place -> organize -> leave.
3. A new, visible, and definite semantic outcome occurs within the same activity, for example: a tool is found and picked up, placement is completed, or an opening/closing action is completed.

Do not create a boundary only because of a small camera rotation within the same region, hand motion, temporary occlusion, illumination changes, or minor visual differences between sampled frames. Motion of the camera or wearer is not an object position_change.

Region definitions (`region`). Select exactly one fixed enum value. Use `other_area` when the region cannot be assigned reliably:
- `sink_area`: washing, rinsing, filling, or draining around the sink
- `food_prep_area`: cutting, peeling, food preparation, or plating around a worktop
- `stove_area`: frying, boiling, heating, or other cooking around a stove
- `refrigerator_area`: retrieving food from or placing food into a refrigerator
- `cabinet_area`: retrieving or storing utensils, containers, or condiments in cabinets, drawers, or shelves
- `other_area`: navigation, transitions, or content outside the functional regions above

Segment roles (`segment_role`):
- `observation`: coherent viewing of a region or scene without a dominant interaction
- `interaction`: a semantically coherent manipulation or operation is being performed
- `navigation`: walking, approaching, leaving, or moving between regions
- `unusable`: black frames, severe blur, or content that cannot be interpreted reliably

Activity types (`activity_type`). Select exactly one fixed enum value:
`observe`, `navigate`, `retrieve`, `place`, `clean`, `prepare`, `cook`, `open_close`, `organize`, `other_interaction`, or `unusable`.

Definite change outcomes (`change_types`). This is a JSON array and may contain multiple values or be empty:
- `position_change`: an object is picked up, placed, or moved to a new location
- `state_change`: an object's functional or physical state changes, such as cutting, filling, heating, opening, or closing
- `identity_replacement`: one item is visibly replaced by another item
Occlusion, reappearance, and camera motion are not change outcomes.

Output schema:
[
  {
    "start_sec": <float>,
    "end_sec": <float>,
    "region": "<sink_area/food_prep_area/stove_area/refrigerator_area/cabinet_area/other_area>",
    "segment_role": "<observation/interaction/navigation/unusable>",
    "activity_type": "<one fixed activity type>",
    "change_types": ["<zero or more definite change outcomes>"],
    "description": "<concise English description, at most 150 characters>"
  }
]

Strict constraints:
- Use half-open intervals `[start_sec, end_sec)` and always satisfy `end_sec > start_sec`.
- The first segment must start at {interval_start}, and the final segment must end at {interval_end}.
- Adjacent segments must be contiguous, non-overlapping, gap-free, and ordered by time.
- Start and end times must align with supplied frame timestamps. Only the final `end_sec` may use the explicitly supplied interval endpoint.
- Output exactly the specified fields. Do not add or omit fields, and do not combine enum values into one string.
- `change_types` must be `[]` for `observation`, `navigation`, and `unusable` segments.
- Merge adjacent content when the region and dominant activity remain the same and no new semantic outcome occurs.
- Use only visible evidence. Do not infer unobserved events.
- Write `description` in English and keep it within 150 characters.
- Output only the JSON array, with no Markdown or explanatory text.

Input:
Cover the complete half-open interval [{interval_start}, {interval_end}).
The first start_sec must be {interval_start}, and the final end_sec must be {interval_end}.

[t={timestamp_1}s]
{frame_1}
[t={timestamp_2}s]
{frame_2}
...
\end{stormprompt}

\begin{stormprompt}{Same-Region Revisit Selection}
<segmentation metadata>
Target region: {region}
Source-video duration: {duration_sec} seconds

You are an expert video-interval selector for the RevisitQA embodied revisit benchmark.

Select 3 to 10 same-region visits from the segmentation metadata below and arrange them chronologically into a dense revisit video. The target output duration is approximately 45 seconds and the hard maximum is 60 seconds. Follow this priority order strictly: same region > proximity to 45 seconds > temporal coverage and change diversity.

Definitions:
- A visit must come from an interval marked `visit_candidate=yes`. Both `observation` and `interaction` segments are eligible, and visible actions or changes inside an interaction may remain in the output.
- Consecutive visits should come from temporally separated occurrences of the same region. Unselected content between visits will be replaced by a short black transition.
- Prefer gaps containing definite changes with non-empty `change_types`, but do not sacrifice the 45-second duration target merely to enforce such a change in every gap.
- Report `start_sec` and `end_sec` in the original source-video timeline; assemble visits chronologically with short black transitions into a ~45s episode (1.5x playback) before 1 FPS sampling.

Constraints:
- Source duration of each visit: 3 to 20 seconds.
- Number of visits: 3 to 10.
- Recommended sum of source durations: 65 to 68 seconds. Brief black gap markers separate consecutive visits, and the assembled output should be close to 45 seconds without exceeding 60 seconds.
- Every visit boundary must match a supplied segmentation boundary (`start_sec` or `end_sec`).
- Prioritize an output duration in the 42-to-48-second range. If it is too short, add more valid interaction or observation visits from the same region.
- Never select `navigation`, `unusable`, or `visit_candidate=no` intervals. Once the duration target is satisfied, prefer broader temporal coverage and more diverse definite change types.
- Every visit must remain within the same functional region {region}. The segments below have already been restricted to this region; do not select across regions.

Return only strict JSON, with no Markdown, code fence, or explanatory text:
{
  "visits": [
    {
      "start_sec": <float>,
      "end_sec": <float>,
      "label": "<concise English description of the visible visit>"
    }
  ],
  "hidden_changes": [
    "<summary of hidden content between visit 1 and visit 2>",
    "<summary of hidden content between visit 2 and visit 3>"
  ],
  "rationale": "<one-sentence English selection rationale>"
}

Input segmentation metadata ({num_segments} segments; timestamps are in seconds):
#{segment_index} | {start_sec}-{end_sec}s | {region} |
role={segment_role} | activity={activity_type} |
changes={change_types} | visit_candidate={yes_or_no} |
{description}
...
\end{stormprompt}

\begin{stormprompt}{Temporally Grounded Kitchen QA Generation}
<timestamped revisit frames>
Episode: {episode_id}
Episode duration: {episode_duration}

You are an expert in constructing QA items for a video understanding benchmark. You will directly observe timestamped frames, ordered chronologically, sampled from one complete episode. Using only this visible video evidence, construct four-choice questions that cannot be answered from common sense alone.

Definitions of the six question types:
- `factual_retrieval`: Retrieve an object, attribute, location, or simple relation from one principal interval in the earlier history. Prefer asking after the target has left the current view.
- `current_state`: Integrate all visible history up to `query_time` to answer the object's current location, attribute, or relation at that time. The query should occur after at least one state change.
- `state_change`: Distinguish a change of state from the process that caused the change. If only the before and after states are visible, ask only what state changed into what. Ask how the change occurred only when the sampled frames directly show the changing action itself. Include separate evidence for at least the before and after states.
- `object_tracking`: Label an item known only when stable, visible, and discriminative identity cues persist across time. Category agreement alone never establishes that two observations are the same instance. If the target crosses a black interval, leaves the field of view, or has similar candidates for which replacement cannot be ruled out, label the item uncertain.
- `temporal_reasoning`: Determine the order, first, last, or kth occurrence among two or more observed events. The order must not be inferable from common sense.
- `history_aggregation`: Up to `query_time`, count events, deduplicate items, accumulate a set, or analyze co-occurrences. The question must state the counting unit explicitly.

Return a strict JSON array. Every question must follow this complete schema:
{
  "id": "filled in later by the program",
  "episode_id": "{episode_id}",
  "query_time": 30.0,
  "question_type": "<one of the six enumerated types>",
  "question_subtype": "<concise English snake_case>",
  "video_evidence": "<minimal visible history required for the answer>",
  "question": "<question in English>",
  "options": ["option A", "option B", "option C", "option D"],
  "answer_index": 0,
  "evidence_spans": [[12.0, 14.0]],
  "diagnostics": {
    "epistemic_status": "known|uncertain",
    "uncertainty_sources": []
  },
  "diagnostic_rationale": {
    "volatility": "<state revisions or historical complexity>",
    "uncertainty": "<why the evidence is sufficient or uncertain>"
  }
}

Hard constraints:
1. Every question must have exactly four distinct options, and `answer_index` must be an integer from 0 to 3. The answer must be uniquely supported by the input video frames. Vary the position of the correct answer.
2. `query_time` and both endpoints of every evidence span must be timestamps shown explicitly on input frames. Every evidence span must end at or before `query_time`.
3. Keep `evidence_spans` as short as possible and include only the minimal sufficient evidence. Use `[t, t]` for single-frame evidence. For a state change, prefer separate point evidence for the before and after states. Use `[start, end]` only when multiple consecutive sampled frames jointly support one visible process. Never substitute a broad video interval for precise evidence.
4. `epistemic_status` has two values. `known` means the current evidence supports one unique answer. `uncertain` means the evidence does not determine the answer. For `known`, `uncertainty_sources` must be `[]`. For `uncertain`, `uncertainty_sources` must contain one or more of: `missing_observation`, `partial_observation`, `low_visual_quality`, `ambiguous_evidence`, `ambiguous_attribute`, or `multiple_candidates`.
5. `history_aggregation` may count only visible observation units. State explicitly whether the unit is "distinct visible action segments," "sampled frames containing the target," or "distinguishable instances." Consecutive sampled frames showing the same action count as one visible action segment. Never infer the true number of actions that may have occurred between sampled observations.
6. Do not mistake camera motion for object motion. Do not use a plan, common sense, or off-screen inference to fill an unobserved process.
7. An uncertain item may ask about scene-relevant information that was not recorded, was only partially recorded, or cannot be determined uniquely. Its correct option must explicitly express why the requested fact cannot be established from the available visual evidence, and the item must include valid `uncertainty_sources`. Use concise, question-specific wording and never present absent content as a known fact.
8. The input is a sparse sample from the complete video, not continuous observation. A process between two timestamps was not directly recorded. Do not claim that an action process, causal relation, or identity continuity was fully observed unless frames before, during, and after the change jointly support it.
9. Output JSON only, with no explanation.

Input:
{
  "episode_id": "{episode_id}",
  "episode_duration": {episode_duration},
  "instruction": "The timestamped images below are the complete visible evidence. Use only timestamps explicitly shown with those images."
}

[t={timestamp_1}s]
{frame_1}
[t={timestamp_2}s]
{frame_2}
...
\end{stormprompt}

%% file: contents/appendix_sim_prompt_refinement.tex
\subsection{Simulation template refinement}
\label{app:sim-template-refinement}

The simulation questions originate from fixed programmatic templates. During template development, a VLM assisted in polishing and revising their wording. The editing stage operates on reusable question patterns and their construction rules. The revised patterns are incorporated into the template library, and the exporter instantiates them with episode-specific objects, events, and query boundaries. Programmatic checks then validate the resulting questions, labels, temporal fields, and answer choices. Thus, language refinement precedes deterministic dataset export.

The following consolidated prompt is the development-time editing brief. Braced fields denote inputs supplied for a particular editing pass. The program constructs the answer choices (three substantive alternatives followed by the shared abstention option). Released builders and counts follow Table~\ref{tab:sim-qa-rules} (no Uncertain state-change items); Appendix~\ref{app:sim-thor-paired} and Appendix~\ref{app:sim-virtualhome} give the instantiation rules.

\begin{stormprompt}{Sim QA Template Refinement: Role and Inputs}
You are a VLM assisting with the language refinement of an online-video question-answering benchmark built in AI2-THOR and VirtualHome.

Task
Revise the supplied fixed question templates so that their instantiated questions are natural, precise, and varied. Preserve the task meaning and its grounding in the available visual evidence. Return reusable templates, not answers to benchmark questions.

The simulation pipeline already provides event records and query-time rules. The program will instantiate your templates, construct answer choices, assign answer positions, and run validation. Your role is to improve how the existing questions are expressed.

Inputs
- Existing template library: {template_library}
- Question-family definitions: {family_definitions}
- Allowed placeholders and their meanings: {placeholder_schema}
- Event-selection and query-time rules: {temporal_rules}
- Concrete-option and abstention-option rules: {option_rules}
- Known/Uncertain labels and source-label conventions: {label_rules}
- Representative instantiated QA records: {example_records}
- Findings from previous wording or shortcut checks, if available: {audit_feedback}
- Requested number of variants per family: {variant_counts}

Use the supplied records to understand each template's semantics. Distinguish observations available by the query time from simulator information used internally to construct the item. An internal event or object identifier does not establish what a viewer could infer from the video.

If a required input is absent or inconsistent, identify the affected template and the missing information. Do not invent an object, event, observation, label, or validation result to fill the gap.
\end{stormprompt}

\begin{stormprompt}{Sim QA Template Refinement: Semantic Contract}
Preserve the following simulation question families.

1. current_state
Known: identify the scene or object state visible at the current query point.
Uncertain: ask for a hidden physical state while the decisive view is missing.

2. factual_retrieval
Known: identify earlier scene or object state along the trajectory.
Uncertain: ask for an object's state during an interval without the observations needed to determine that location.

3. history_aggregation
Known: identify the object that returns last in the selected pair of visible return events.
Uncertain: ask how many separate manipulations occur while a named object is out of view.

4. state_change
Known: identify which object undergoes the selected disappearance or reappearance.

5. object_tracking
Known: identify the object category that returns after an earlier disappearance.
Uncertain: ask whether the reappearing object is the same physical instance across an unobserved interval.

6. temporal_reasoning
Known: identify the second object to return in the selected pair of return events.
Uncertain: ask for the order of actions that occur outside the available observation.

These definitions are fixed for this editing pass. In particular, do not turn room identification into object-state recognition, a last-return question into a count question, or category-level return into a claim about physical identity. Preserve the distinction between visible state change and an unobserved explanation of that change.

Editable content
- Wording, grammar, sentence structure, and natural temporal expressions.
- Semantically equivalent variants of a supplied question pattern.
- Option phrasing only when the supplied option rules explicitly allow it and the option's meaning remains unchanged.

Fixed content
- Question type and the semantic operation being evaluated.
- Event-selection rule, query_time, and evidence_spans.
- Target bindings and the correct semantic answer.
- Known/Uncertain status and uncertainty_sources.
- Number and meanings of answer alternatives.
- Dataset quotas and the program's answer-position assignment.

For concrete room or object-category alternatives, retain the supplied category names. Do not rewrite an Uncertain item as a question with a visually established concrete answer.
\end{stormprompt}

\begin{stormprompt}{Sim QA Template Refinement: Evidence and Language Rules}
Temporal grounding
- Interpret every question relative to its existing query boundary.
- Use only relations licensed by the selected event history.
- Preserve whether the question concerns the current view, an earlier observation, a selected local sequence, or an unobserved gap.
- Use expressions such as "at the current query point", "after disappearing earlier", or "within the final pair of visible return events" only when they match the supplied event-selection rule.
- Do not introduce an observation after query_time. Do not convert an anticipated return into an event already observed by the query.
- Keep absolute times in the structured query_time and evidence_spans fields. Do not place seconds, timestamps, or numeric characters in the rendered question text.
- Do not interpret an evidence interval as uninterrupted visibility: it may span a gap in which the target is not observed.

Evidence and uncertainty
- Describe visible appearance, absence, return, or order without inventing a hidden physical action.
- Do not treat camera rotation as object motion.
- Do not infer carrying, rotation, destruction, replacement, or identity continuity from disappearance and reappearance alone.
- For Uncertain questions, preserve the requested hidden fact and the existing source labels. Do not resolve the question using privileged simulator records.
- Do not add or remove uncertainty_sources based on stylistic preference. Flag a suspected annotation inconsistency for separate review.

Natural wording
- Write one direct question about one existing semantic target.
- Use ordinary descriptions of objects, rooms, and visible events.
- Remove code-oriented language, opaque identifiers, and awkward references to an "indexed transition" when a precise natural temporal expression conveys the same relation.
- Avoid leading wording such as "the correct answer", "the known object", "uniquely supported", or "the evidence proves" in the question stem.
- Avoid unnecessary verbosity, repeated qualifications, and unsupported details about color, material, position, or appearance.
- Keep the question understandable without access to the annotation files.

Shortcut control
- For Known questions that ask which object or room is present, do not reveal the correct category in the stem.
- A named object may appear when it is the fixed target of a hidden-location, hidden-action, or identity question; its name must not determine the answer.
- Do not insert scene names, episode identifiers, seeds, answer letters, or arbitrary numbers to make questions look different.
- Where option wording is editable, keep alternatives comparable in grammar, specificity, and descriptive detail. Do not give only the correct alternative an explanation or a lexical match to the stem.
- Keep stems free of answer cues that can be resolved from text alone.

Diversity
- Produce the requested variants within each semantic family.
- Vary sentence structure and natural temporal phrasing while preserving the same answer rule.
- Avoid exact duplicates and meaningless changes in punctuation.
- Do not create diversity by changing the number of events, temporal scope, queried attribute, or required reasoning operation.
- If the requested diversity cannot be obtained without changing meaning, flag the affected family rather than silently changing the task.
\end{stormprompt}

\begin{stormprompt}{Sim QA Template Refinement: Output and Self-Check}
Return JSON with the following structure. Include one family entry for each supplied question_type and epistemic_status combination. Preserve any additional distinctions required by the input, such as disappearance versus reappearance.

{
  "template_families": [
    {
      "question_type": "{existing_question_type}",
      "epistemic_status": "{existing_status}",
      "answer_rule_ref": "{existing_rule_identifier}",
      "variants": [
        {
          "variant_id": "{internal_variant_identifier}",
          "question_template": "{revised_question_with_allowed_placeholders}",
          "required_bindings": ["{allowed_placeholder_name}"],
          "temporal_anchor": "{existing_query_or_event_relation}",
          "edit_note": "{brief_description_of_the_language_change}"
        }
      ]
    }
  ],
  "needs_review": [
    {
      "template_ref": "{affected_input_template}",
      "issue": "{missing_input_or_semantic_conflict}",
      "required_decision": "{information_needed_to_resolve_it}"
    }
  ]
}

Use an empty needs_review array when no issue remains. Keep internal identifiers and edit notes outside the rendered question. Do not output newly assigned answers, labels, timestamps, or release statistics.

Before returning the JSON:
1. Check each variant against its original answer rule.
2. Check that every placeholder is allowed and has a defined binding.
3. Check that the temporal relation is valid at the existing query time.
4. Check that Known wording contains no answer cue or added abstention, and that Uncertain wording preserves the unresolved fact.
5. Check that the instantiated question will contain no numeric timestamp, opaque identifier, or unsupported visual detail.
6. Check the variant set for duplicate text and changes in task meaning.
7. Put issues requiring new evidence or a changed construction rule in needs_review instead of resolving them through wording.
\end{stormprompt}

\paragraph{VHome template constraints.}
The VHome builders use the same meaning-preserving editing principle, with three additional constraints: keep the explicit query-time restriction, distinguish visible absence from unknown physical state, and retain the common abstention choice on every item. Return/exit wording must preserve the two-consecutive-sample rule. The final deterministic language pass leaves all evidence intervals, query times, options' order, answer indices, and epistemic labels unchanged (Appendix~\ref{app:sim-virtualhome}).